\documentclass[lettersize,journal]{IEEEtran}
\usepackage{amsmath,amsfonts}
\usepackage{algorithmicx}
\usepackage{array}
\usepackage{textcomp}
\usepackage{stfloats}
\usepackage{url}
\usepackage{verbatim}
\usepackage{graphicx}
\def\BibTeX{{\rm B\kern-.05em{\sc i\kern-.025em b}\kern-.08em
    T\kern-.1667em\lower.7ex\hbox{E}\kern-.125emX}}
\usepackage{balance}
\usepackage{cite}
\usepackage{enumitem}
\usepackage{algorithm}
\usepackage{algpseudocode}
\usepackage{booktabs}
\usepackage{multirow}
\usepackage{subfigure}
\usepackage[colorlinks,
linkcolor=black,
anchorcolor=black,
citecolor=black,
urlcolor=black,
]{hyperref}
\usepackage{xcolor}
\definecolor{g}{RGB}{83, 120, 3}
\begin{document}
\title{HTD-Mamba: Efficient Hyperspectral Target Detection with Pyramid State Space Model}
	\author{Dunbin~Shen,
	Xuanbing~Zhu,
	Jiacheng~Tian,
	Jianjun~Liu,~\IEEEmembership{Member,~IEEE},
	Zhenrong~Du,
	Hongyu~Wang,~\IEEEmembership{Member,~IEEE},
	Xiaorui~Ma,~\IEEEmembership{Member,~IEEE}
	\thanks{This work was supported in part by the National Natural Science Foundation of China under Grant 62271102, in part by the Fundamental Research Funds for the Central Universities under Grant DUT23YG117, in part by the Dalian Science and Technology Innovation Foundation under Grant 2022JJ11CG002, and in part by the Liaoning Province Science and Technology Planning Project under Grant 2022JH1/10800100. \textit{(Corresponding author: Xiaorui~Ma.)}
	}
	\thanks{Dunbin Shen, Xuanbing Zhu, Jiacheng Tian, Zhenrong Du, Hongyu Wang, and Xiaorui Ma are with the School of Information and Communication Engineering, DUT Artificial Intelligence Institute, and the Dalian Key Laboratory of Artificial Intelligence at Dalian University of Technology, Dalian 116024, China (e-mail: sdb\_2012@163.com;  zhuxuanbing@mail.dlut.edu.cn; tjcheng@mail.dlut.edu.cn; duzhenrong@dlut.edu.cn; whyu@dlut.edu.cn; maxr@dlut.edu.cn).}
	\thanks{Jianjun Liu is with the Jiangsu Provincial Engineering Laboratory for Pattern Recognition and Computational Intelligence, Jiangnan University, Wuxi 214122, China (e-mail: jianjunliu@jiangnan.edu.cn).}
}

\markboth{Journal of \LaTeX\ Class Files,~Vol.~xx, No.~xx, July~2024}%
{How to Use the IEEEtran \LaTeX \ Templates}

\maketitle

\begin{abstract}
	Hyperspectral target detection (HTD) identifies objects of interest from complex backgrounds at the pixel level, playing a vital role in Earth observation. However, HTD faces challenges due to limited prior knowledge and spectral variation, leading to underfitting models and unreliable performance. To address these challenges, this paper proposes an efficient self-supervised HTD method with a pyramid state space model (SSM), named HTD-Mamba, which employs spectrally contrastive learning to distinguish between target and background based on the similarity measurement of intrinsic features. Specifically, to obtain sufficient training samples and leverage spatial contextual information, we propose a spatial-encoded spectral augmentation technique that encodes all surrounding pixels within a patch into a transformed view of the center pixel. Additionally, to explore global band correlations, we divide pixels into continuous group-wise spectral embeddings and introduce Mamba to HTD for the first time to model long-range dependencies of the spectral sequence with linear complexity. Furthermore, to alleviate spectral variation and enhance robust representation, we propose a pyramid SSM as a backbone to capture and fuse multiresolution spectral-wise intrinsic features. Extensive experiments conducted on four public datasets demonstrate that the proposed method outperforms state-of-the-art methods in both quantitative and qualitative evaluations. Code is available at \url{https://github.com/shendb2022/HTD-Mamba}.
\end{abstract}


\section{Introduction}
\IEEEPARstart{H}{yperspectral} imaging sensors capture contiguous spectral profiles of surface materials, spanning from the ultraviolet to the shortwave infrared bands. By scanning across a scene, each spectral narrowband yields a two-dimensional (2D) geometric image. The composite of these images forms a three-dimensional (3D) data cube, known as the hyperspectral image (HSI). In this cube, the two spatial dimensions depict surface structure, while the third dimension characterizes spectral information. The distinct spectral signatures, arising from variations in absorption, reflection, and scattering across different materials, serve as unique identifiers akin to fingerprints for precise recognition of ground objects\cite{plaza2009recent}. As a result, HSIs are extensively used in applications such as military reconnaissance\cite{shimoni2019hyperspectral}, medical diagnostics\cite{karim2023hyperspectral}, ecological monitoring\cite{marconi2022continental}, and precision agriculture\cite{pande2023application}.

Hyperspectral target detection (HTD), a critical downstream task, aims to identify and locate objects of interest based on spectral and spatial discrepancies. Unlike typical “box” detection methods used in RGB images, which delineate objects with various rectangular or rotated boxes, HTD distinguishes between background and target information at the pixel or sub-pixel level. This fine-grained detection and joint spectral-spatial discrimination make HTD highly relevant in current Earth observation and quantitative remote sensing tasks\cite{chen2023target, sneha2022hyperspectral, liu2023spectrum}. However, it faces two longstanding challenges:
\begin{enumerate}[label=\textbullet]
	\item Limited prior knowledge: The task requires that only a few target spectra (usually one) can be used as supervised information for HTD, leading to model underfitting and making it difficult to optimize.
	\item Spectral variation: The same material can exhibit diverse spectra, and different materials can appear with similar spectral profiles, making it difficult to achieve reliable and robust performance based solely on spectral discrepancy.
\end{enumerate}

Over the past decades, various approaches have been developed to address these challenges. Early statistical-based methods \cite{kraut2005adaptive, chang2005orthogonal, broadwater2007hybrid, vincent2020one, chen2022glrt} assumed certain background distributions to determine target pixels. However, their performance may be limited when applied to real-world datasets due to inconsistencies with prior assumptions. Subsequent representation-based approaches \cite{zhang2014sparse, li2015combined, cheng2021decomposition, chen2022background, zeng2022sparse} transformed the detection task into an optimization problem based on the linear mixing model and various regularizations. Despite their physical interpretability, these models may encounter difficulties due to inaccurate regularization constraints and weak nonlinear representations.

In contrast, deep learning, with its powerful feature extraction and parallel computation capabilities, has demonstrated effectiveness and superiority in HTD. Due to limited prior knowledge and class imbalance between targets and backgrounds, two categories of deep detectors have emerged: supervised classification methods \cite{zhang2020htd, zhu2020two, wang2022meta, zhu2022target, jiao2023triplet} and unsupervised detection methods \cite{shi2019discriminative, xie2020background, li2022target, wang2023self, tian2024hyperspectral}. The former expands the training samples via data augmentation and employs a similarity-dissimilarity binary classification model for detection. The latter extracts discriminative intrinsic features through a carefully designed self-supervised model and then separates the target from the background based on these features. Although these methods have achieved considerable performance, several problems still exist. Firstly, it is challenging for supervised classification methods to train an accurate binary classifier due to the distribution mismatch between the expanded samples and the original samples. Secondly, it is difficult for unsupervised detection methods to design a discrimination loss that ensures effective background-target separation. Thirdly, it is demanding for both categories of methods to develop an efficient backbone to obtain robust intrinsic feature representations for HTD.

Current backbones for HTD are mainly based on convolutional neural networks (CNNs) and Transformers. CNN-based methods \cite{zhang2020htd, zhu2020two, wang2022meta, wang2023self} excel at extracting local features but suffer from a limited receptive field, making it difficult to capture non-local correlations. In contrast, Transformer-based methods \cite{jiao2023triplet, rao2022siamese, girard2022swin, li2023htdformer} can capture global contextual dependencies based on the multihead self-attention mechanism. However, the computational complexity of Transformers is quadratic to the number of embedding tokens, leading to substantial computational overload and memory requirements as the sequence grows. Recently, structured state space models (SSMs) \cite{gu2021efficiently} have gained attention for modeling long sequences with near-linear complexity. Notably, Mamba \cite{gu2023mamba} introduces a selective state space (S6) that makes SSM parameters input-aware, allowing for the selective propagation or forgetting of information based on the current token. Additionally, a hardware parallel scan algorithm in recurrent mode accelerates the training and inference process. Therefore, considering the high-dimensional sequence nature of hyperspectral data, applying Mamba as a backbone for HTD shows promise in efficiently capturing long-range dependencies.

Inspired by the above insights, we propose an efficient HTD method with a pyramid SSM (HTD-Mamba). This method employs spectrally contrastive learning to recognize pixel-wise instances by maximizing the feature similarity between each pixel and its transformed view. Thus, detection can be performed by feeding pairs consisting of each detected pixel and the prior target spectrum into the well-trained model. Specifically, an efficient data augmentation technique is proposed to obtain sufficient training samples and leverage spatial information. This technique encodes all surrounding pixels within a patch into a new spectral view of the center pixel, using spectral similarity as weights. Additionally, to explore global band correlations, the input pixels are dynamically divided into continuous group-wise spectral embeddings using a one-dimensional (1D) CNN. The efficient architecture, Mamba, is first introduced to HTD to model long-range dependencies of the spectral sequence with linear complexity. Furthermore, to enhance robust feature representations, a pyramid SSM is developed as a backbone to extract multiresolution spectral-wise intrinsic features by capturing the global correlation of spectral sequences at different resolutions and fusing features from different levels. Therefore, HTD-Mamba effectively addresses the challenges of limited prior knowledge by constructing sufficient view pairs and mitigates spectral variation by extracting multiresolution discriminative intrinsic features using the pyramid SSM. Experimental results across multiple datasets demonstrate the effectiveness and superiority of HTD-Mamba compared to existing state-of-the-art methods.

In summary, the main contributions can be outlined as follows:
\begin{enumerate}[label=\textbullet]
	\item We propose a self-supervised spectrally contrastive learning method based on Mamba to address the issues of limited prior knowledge and spectral variation, which recognize pixel-wise instances by maximizing the feature similarity between each pixel and its transformed view. To the best of our knowledge, we are the first to introduce Mamba to HTD, capturing long-range dependencies of the spectral sequence with linear complexity.
	\item We propose an efficient data augmentation technique to create sufficient view pairs and leverage spatial contextual information, which encodes all surrounding pixels within a patch into a transformed view of the center pixel.
	\item We propose a pyramid SSM to alleviate spectral variation and enhance robust representation, which extracts multiresolution spectral-wise intrinsic features by capturing the global correlation of spectral sequences at different resolutions and fusing features from different levels.
\end{enumerate}

The rest of the paper is organized as follows. Section \ref{methodology} introduces the preliminary concepts behind Mamba and details our proposed approach. Section \ref{experiments} presents the experimental results and analysis. Finally, Section \ref{conclusion} concludes the paper.  
\section{Methodology}
\label{methodology}
In this section, we first introduce the basic concepts of Mamba, including SSM, discretization, and the selective scan mechanism. Subsequently, we provide a detailed explanation of the proposed method, covering the overall architecture, module design, and loss function.
\subsection{Preliminary}
\subsubsection{SSM}
The SSM is known as a linear time-invariant system that maps a 1D function or sequence $x(t) \in \mathbb{R}$ to its response $y(t)\in \mathbb{R}$. Mathematically, this process can be formulated using the following linear ordinary differential equations (ODEs):
\begin{equation}
	\begin{split}
		h'(t) &= {\bf A}h(t) + {\bf B}x(t), \\
		y(t) &= {\bf C}h(t)
	\end{split}
	\label{ssm}
\end{equation}
where $h(t)\in \mathbb{R}^{D}$ denotes the hidden state with size $D$,  $h'(t)$ denotes the  time derivative of  $h(t)$, ${\bf A} \in \mathbb{R}^{D \times D}$ is the state transition matrix derived from the high-order polynomial projection operator to retain historical information and capture long-range dependencies, and ${\bf B} \in \mathbb{R}^{D \times 1}$ and  ${\bf C} \in \mathbb{R}^{1 \times D}$ are the projection matrices controlling the input and output of the system, respectively.
\begin{figure*}[!t]
	\centering
	\includegraphics[width= 1.0 \textwidth]{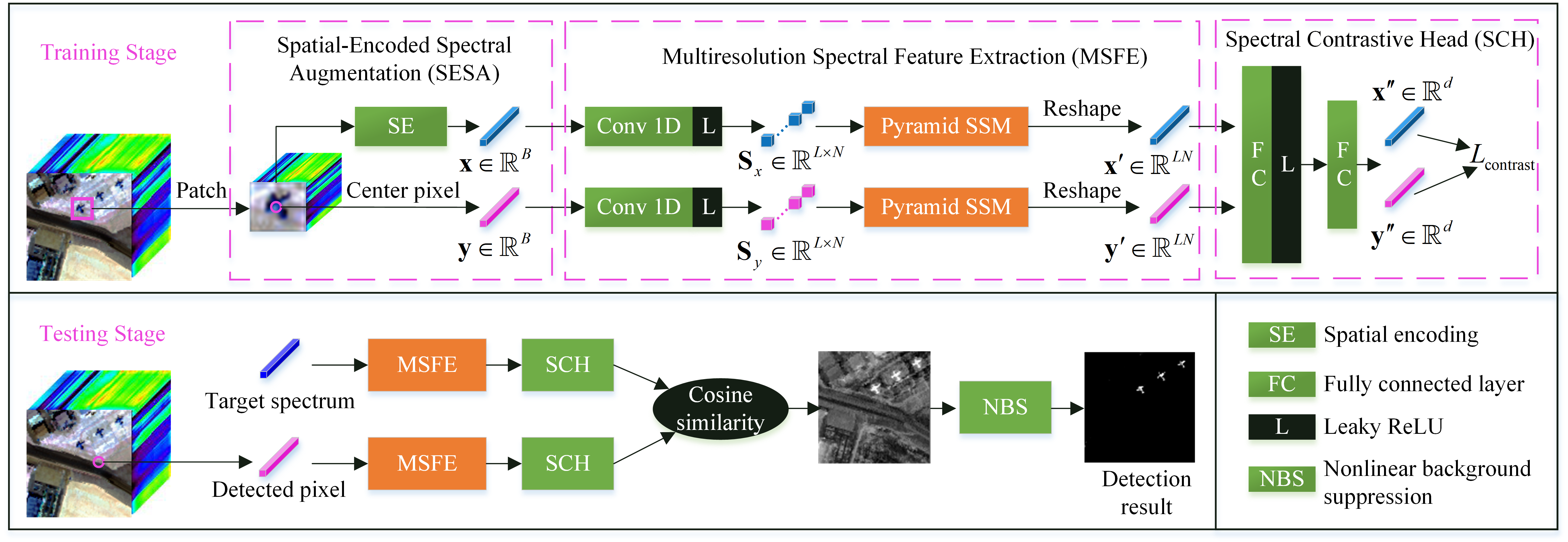}
	\caption{The overall architecture of the proposed HTD-Mamba.}
	\label{architecture}
\end{figure*}
\begin{figure*}[!t]
	\centering
	\includegraphics[width= 0.8 \textwidth]{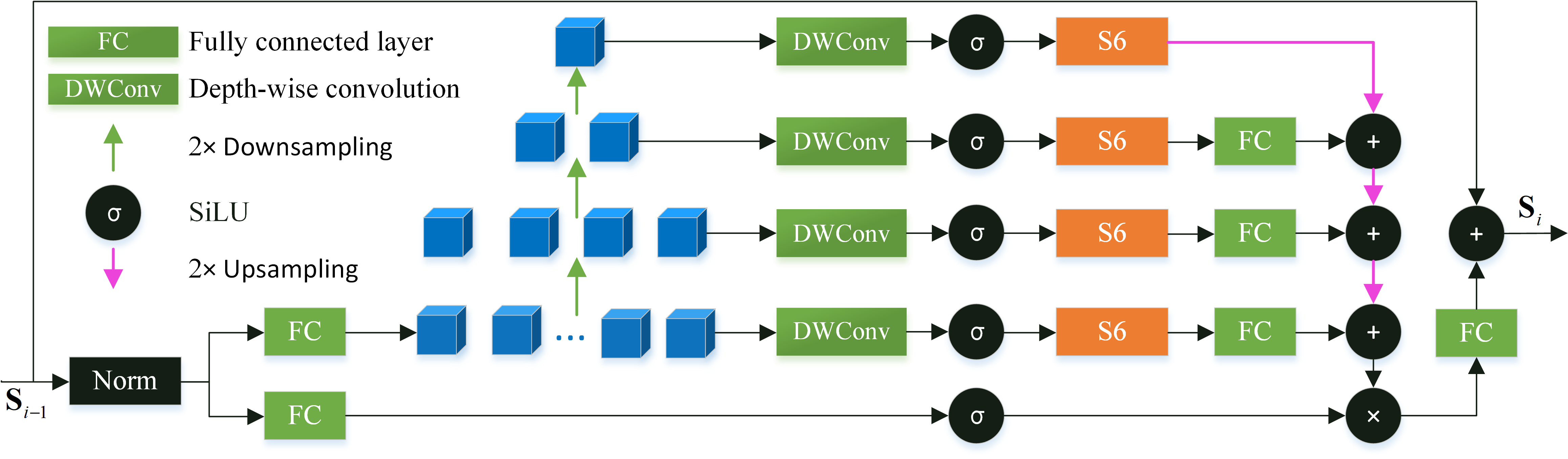}
	\caption{Flowchart of pyramid SSM.}
	\label{pyramid_ssm}
\end{figure*}
\subsubsection{Discretization} To integrate the continuous SSM into deep learning algorithms, the discretization process is necessary. Specifically, the ODEs \eqref{ssm} can be discretized using a zero-order hold rule as follows:
\begin{equation}
	\begin{split}
		h_t&= \bar{{\bf A}}h_{t-1} + \bar{{\bf B}}x_t, \\
		y_t&= {\bf C}h_t
	\end{split}
	\label{discretized_ssm}
\end{equation}
where 
\begin{equation}
	\begin{split}
		\bar{{\bf A}}&= {\rm exp} ({\bf \Delta}{\bf A}), \\
		\bar{{\bf B}}&= ({\bf \Delta}{\bf A})^{-1}({\rm exp}({\bf \Delta}{\bf A} - {\bf I}))\cdot{\bf \Delta}{\bf B}
	\end{split}
	\label{discretized_matrix}
\end{equation}
and ${\bf \Delta}$ is a timescale parameter that balances the influence of the current input and the preceding state. In practice, $\bar{{\bf B}}$ is approximated as ${\bf \Delta}{\bf B}$, follwing the first-order Taylor series.
\subsubsection{Selective Scan Mechanism}
Due to the time-invariant property, i.e., $\bf B$, $\bf C$, and ${\bf \Delta}$ are independent of the input sequence, the output can be calculated either through recurrance or by global convolution. However, this approach limits context awareness. To address this limitation, a selective scan mechanism is incorporated, where the projection matrices and time scale are derived from the input, ensuring awareness of the contextual information:
\begin{equation}
	\begin{split}
		S_{\bf B}({\bf z})&= {\rm Linear}_N({\bf z}), \\
		S_{\bf C}({\bf z})&= {\rm Linear}_N({\bf z}), \\
		S_{\bf \Delta}({\bf z})& = {\rm Broadcast}_D({\rm Linear}_1({\bf z}))
	\end{split}
	\label{input_dependent}
\end{equation}
where ${\rm Linear}_q$ represents the parameterized projection to dimension $q$, ${\rm Broadcast}_q$ denotes  the broadcast operation to dimension $q$, and $N$ is the embedding size. Thus, the system becomes time-varying, and only recurrence can be adopted. Additionally, with the incorporation of parallel scanning and integration with the gated multilayer perceptron (MLP), the SSM evolves into the popular Mamba model, which excels at capturing long-range dependencies with near-linear time complexity and hardware acceleration.
\subsection{Overall Architecture}
To address the issues of limited training samples and spectral variation, we propose a self-supervised spectrally contrastive learning method based on Mamba with a pyramid SSM. This method aims to identify pixel-wise instances by maximizing the similarity and consistency between each pixel and its transformed view. Thus, targets of interest can be detected by pairing them with the prior target spectrum. Fig. \ref{architecture} illustrates the overall architecture of the proposed method.

The training stage encompasses spatial-encoded spectral augmentation (SESA), multiresolution spectral feature extraction (MSFE), and spectral contrastive head (SCH). The SESA block initially creates a new spectral view of the center pixel by encoding contextual information within a patch, thereby generating sufficient view pairs for contrastive learning. Subsequently, the MSFE block divides the pixel into continuous group-wise spectral embeddings, treating them as a spectral sequence. To achieve discriminative intrinsic feature representation, we propose a pyramid SSM that captures multiresolution long-range dependencies within sequences. Finally, the paired extracted features pass through the SCH block to compute similarity, and contrastive loss is applied to ensure model discrimination by pulling positive pairs closer while pushing negative pairs farther apart.

In the testing stage, each detected pixel is paired with the prior target spectrum and then passes through the MSFE and SCH blocks to obtain the discriminative intrinsic feature vectors. The detection result of the $i$-th pixel is obtained by calculating the cosine similarity:
\begin{equation}
	\mu_i = c(f({\bf X}_i), f({\bf d}))
\end{equation}
where $c({\bf a}, {\bf b})=({\bf a}^T{\bf b})/(\|{\bf a}\|_2\cdot \|{\bf b}\|_2)$ denotes the cosine similarity function,  $\|\cdot\|_2$ denotes the L2 norm, ${\bf X}\in \mathbb{R}^{n\times B}$ denotes the image matrix with $n$ pixels and $B$ bands, ${\bf d} \in \mathbb{R}^{B}$ denotes the target spectrum, and $f$ denotes the backbone consisting of MSFE and SCH. Finally, a nonlinear background suppression (NBS) function is applied to suppress the background:
\begin{equation}
	\mu'_i = {\rm exp}(-\frac{(\mu_i-1)^2}{\delta})
\end{equation}
where $\delta>0$ is a tunable parameter that controls the degree of background suppression.
\subsection{Spatial-Encoded Spectral Augmentation}
Based on the fact that the performance of contrastive learning relies heavily on appropriate data augmentation, we propose a simple yet efficient spectral augmentation technique that encodes the spatial context within a patch into a new spectral view of the center pixel. This technique aims to leverage local spatial information and provide sufficient view pairs for contrastive learning.

Given a random patch ${\bf P} \in \mathbb{R}^{p^2 \times B}$, where the patch size is $p\times p$, and the center pixel is ${\bf y}  \in \mathbb{R}^{B}$. By encoding all pixels within the patch using spectral similarity, the new spectral view $\bf x$ can be expressed as 
\begin{equation}
	{\bf x} = {\bf P}^T{\bf W}
\end{equation}
where ${\bf W} \in \mathbb{R}^{p^2 \times 1}$ is the weight matrix measuring the similarity between the center pixel and each pixel within the patch. The similarity weight corresponding to the $i$-th pixel within the patch is defined as
\begin{equation}
	{\bf W}_i = \frac{{\rm exp}(c({\bf y},{\bf P}_i))}{\sum\limits_{j=1}^{p^2}{\rm exp}(c({\bf y},{\bf P}_j))}.
\end{equation}
After obtaining the new spectral view of the center pixel, pairs $({\bf x}, {\bf y})$ from the same location comprise the positive samples while pairs from different locations comprise the negative samples. As a result, sufficient training samples can be obtained for contrastive learning.

\subsection{Multiresolution Spectral Feature Extraction}
To mine discriminative intrinsic features from different spectral views, we treat the group-wise spectral embeddings as a sequence and propose a pyramid SSM to capture multiresolution long-range spectral dependencies.

\emph{1) Group-Wise Spectral Embedding:} The adjacent continuous bands of the hyperspectral spectrum provide detailed information for object recognition. Therefore, we adopt a learnable embedding module to convert the input spectrum into group-wise tokens.

Specifically, we use 1D convolution with a  stride to extract local features and divide the spectrum into group representations:
\begin{equation}
	{\bf S}_x = \sigma_1({{\rm Conv}({\bf x}, m, s, N)})
\end{equation}
where ${\rm Conv}$ denotes the 1D convolution, $m$ denotes the kernel size, $s={\rm max}(1, \lfloor m/4\rfloor)$ denotes the stride, $N$ denotes the number of kernels, and $\sigma_1$ is the leaky rectified linear unit (Leaky ReLU)\cite{xu2015empirical} function for nonlinear representation. After this process, paired spectral samples $({\bf x}, {\bf y})$ can be converted to group-wise embeddings $({\bf S}_x \in \mathbb{R}^{L \times N}, {\bf S}_y \in \mathbb{R}^{L \times N})$, where $L$ denotes the number of spectral groups, satisfying $L =\lfloor(B-m) / s\rfloor + 1$. Thus, numerous spectral tokens can be obtained to extract discriminative intrinsic features.

\emph{2) Pyramid SSM:} The resulting group-wise spectral embeddings can be viewed as a spectral sequence. Considering the linear complexity of SSM in capturing long-range dependencies and the benefits of multiresolution analysis, we propose a pyramid SSM to extract multiresolution spectral-wise intrinsic features.

The flowchart of the proposed pyramid SSM is shown in Fig. \ref{pyramid_ssm}. It comprises bottom-up downsampling, global spectral feature extraction, top-down upsampling and feature fusion, and the gated block. In each deep layer, the input sequence first undergoes a normalization operation to reduce the internal covariate shift and stabilize the training process. In this work, the root mean square layer normalization (RMSNorm)\cite{zhang2019root} is applied to improve the computational efficiency by eliminating the mean calculation required in the layer normalization:
\begin{equation}
	{\bar {\bf S}}_{i-1}^j = \frac{{\bf S}_{i-1}^j}{{\rm RMS}({\bf S}_{i-1})}{\bf g}_{i-1}^j
\end{equation}
where ${\rm RMS}$ denotes the root mean square function, ${\bf S}_{i-1}^j$ denotes the $j$-th channel of the $i$-th layer input sequence, and ${\bf g}$ denotes the learnable weight. Then, the normalized sequence is projected into two representations for feature extraction and gated operation, each using a fully connected (FC) layer:
\begin{equation}
	\begin{split}
		{\bf Z}_1 &= {\rm Linear}_{2N}({\bar {\bf S}}_{i-1}), \\
		{\bf Z}_2 &= {\rm Linear}_{2N}({\bar {\bf S}}_{i-1}).
	\end{split}
\end{equation}

In the bottom-up downsampling process, the spectral sequence is gradually downsampled to obtain multiresolution representations, which facilitates comprehensive analysis. Specifically, the adjacent hierarchical sampling is performed using convolution with stride:
\begin{equation}
	{\bf Z}_1^{k} = {\rm Conv}({\bf Z}_1^{k-1}, 3, 2, 2^{k+1}N)
\end{equation}
where ${\bf Z}_1^{0}={\bf Z}_1$, $k \in \{1, 2, 3\}$ denotes the $k$-th pyramid level. The kernel size is fixed at 3, the stride is 2 to perform 2$\times$ downsampling, and the token size is doubled to maintain information capacity.

For each pyramid level, the spectral sequence first passes through a 1D depth-wise convolution with activation to extract local features and then undergoes an S6 module to capture long-range dependencies among tokens. This process can be expressed as
\begin{equation}
	{\bar {\bf Z}_1^{k}} = {\rm S6}(\sigma_2({\rm DWConv}( {\bf Z}_1^{k}, 3, 2^{k+1}N)))
\end{equation}
where $k \in \{0, 1, 2, 3\}$, ${\rm DWConv}$ denotes the 1D depth-wise convolution,  the kernel size is 3, the number of groups is $2^{k+1}N$, $\sigma_2$ denotes the sigmoid-weighted linear unit (SiLU)\cite{elfwing2018sigmoid} function, and padding is used in the convolution to keep the size fixed.

In the top-down upsampling and feature fusion process, low-level features are recovered from the adjacent high level and then fused with the extracted global features from the same level. This process can be expressed as
\begin{equation}
	\begin{split}
		{\hat {\bf Z}_1^{k-1}} &= {\rm DConv}({\hat {\bf Z}_1^{k}}, 3, 2, 2^{k}N)), \\
		{\hat {\bf Z}_1^{k-1}} &= {\hat {\bf Z}_1^{k-1}} + {\rm Linear}_{2^kN}({\bar {\bf Z}_1^{k-1}})
	\end{split}
\end{equation}
where $k \in \{3, 2, 1\}$, ${\hat {\bf Z}_1^{3}} = {\bar {\bf Z}_1^{3}}$, ${\rm DConv}$ denotes the transposed 1D convolution with a kernel size of 3 and a stride of 2 to perform $2\times$ upsampling. The token size is halved to maintain information capacity. The FC layer is used to perform flexible linear transformation in preparation for fusion.
\begin{algorithm}[!t]
	\caption{S6}
	\begin{algorithmic}[1]
		\item[]\hskip-\algorithmicindent \textbf{Require}: Spectral sequence ${\bf z}$: {\color{g} $(P, L, N)$} 
		\item[]\hskip-\algorithmicindent  \textbf{Ensure}:  Spectral sequence $\bar{{\bf z}}$: {\color{g} $(P, L, N)$} 
		\State {\bf B}: {\color{g} $(P, L, D)$} $\leftarrow$ $S_{\bf B}({\bf z})$
		\State {\bf C}: {\color{g} $(P, L, D)$} $\leftarrow$ $S_{\bf C}({\bf z})$
		\State ${\bf \Delta}$: {\color{g} $(P, L, N)$} $\leftarrow$ ${\rm log}(1 + {\rm exp}(S_{\bf \Delta}({\bf z}) + {\rm Parameter}^{\bf \Delta}))$
		\State ${\bar{\bf A}}$: {\color{g} $(P, L, N, D)$} $\leftarrow$ ${\bf \Delta}\otimes{\rm Parameter}^{\bf A}$
		\State ${\bar{\bf B}}$: {\color{g} $(P, L, N, D)$} $\leftarrow$ ${\bf \Delta}\otimes{\bf B}$
		\State $\bar{{\bf z}}$: {\color{g} $(P, L, N)$} $\leftarrow$ ${\rm SSM}({\bar{\bf A}}, {\bar{\bf B}}, {\bf C})({\bf z})$
		\Statex ${\rm Return}: {\bar{{\bf z}}}$
	\end{algorithmic}
	\label{alg_Algorithm1}
\end{algorithm}
\begin{algorithm}[!t]
	\caption{Pyramid SSM}
	\begin{algorithmic}[1]
		\item[]\hskip-\algorithmicindent \textbf{Require}: Spectral sequence ${\bf S}_{i-1}$: {\color{g} $(P, L, N)$} 
		\item[]\hskip-\algorithmicindent  \textbf{Ensure}:  Spectral sequence ${\bf S}_{i}$: {\color{g} $(P, L, N)$}
		\State {\color{gray}/* Normalization and projection */}
		\State  ${\bar {\bf S}}_{i-1}$:  {\color{g} $(P, L, N)$} $\leftarrow$ ${\rm RMSNorm}({\bf S}_{i-1})$
		\State ${\bf Z}_1$: {\color{g} $(P, L, 2N)$}  $\leftarrow$ ${\rm Linear}_{2N}({\bar {\bf S}}_{i-1})$
		\State ${\bf Z}_2$: {\color{g} $(P, L, 2N)$}  $\leftarrow$ ${\rm Linear}_{2N}({\bar {\bf S}}_{i-1})$
		\State {\color{gray}/* Downsampling and feature extraction */}
		\State $L_{0}=L$, ${\bf Z}_1^{0}={\bf Z}_1$
		\State ${\bar {\bf Z}_1^{0}}$: {\color{g} $(P, L_{0}, 2N)$} $\leftarrow$ ${\rm S6}(\sigma_2({\rm DWConv}( {\bf Z}_1^{0})))$
		\For{$k=1$ to 3}
		\State  $L_{k}=L_{k-1}//2$
		\State ${\bf Z}_1^{k}$: {\color{g} $(P, L_{k}, 2^{k+1}N)$} $\leftarrow$ ${\rm Conv}({\bf Z}_1^{k-1})$
		\State ${\bar {\bf Z}_1^{k}}$: {\color{g} $(P, L_{k}, 2^{k+1}N)$} $\leftarrow$ ${\rm S6}(\sigma_2({\rm DWConv}( {\bf Z}_1^{k})))$
		\EndFor
		\State {\color{gray}/* Upsampling and feature fusion */}
		\State ${\hat {\bf Z}_1^{3}}={\bar {\bf Z}_1^{3}}$
		\For{$k=3$ to 1}
		\State ${\hat {\bf Z}_1^{k-1}}$: {\color{g} $(P, L_{k-1}, 2^kN)$} $\leftarrow$ ${\rm DConv}({\hat {\bf Z}_1^{k}}))$ 
		\State ${\hat {\bf Z}_1^{k-1}}$: {\color{g} $(P, L_{k-1}, 2^kN)$} $\leftarrow$ ${\hat {\bf Z}_1^{k-1}} + {\rm Linear}_{2^kN}({\bar {\bf Z}_1^{k-1}})$
		\EndFor
		\State {\color{gray}/* Gated block */}
		\State ${\bf Z}$: {\color{g} $(P, L, 2N)$} $\leftarrow$ ${\hat {\bf Z}_1^{0}} \odot \sigma_2({\bf Z}_2)$
		\State {\color{gray}/* Residual connection */}
		\State ${\bf S}_i$: {\color{g} $(P, L, N)$} $\leftarrow$ ${\bf S}_{i-1} + {\rm Linear}_{N}({\bf Z})$
		\Statex ${\rm Return}: {\bf S}_{i}$
	\end{algorithmic}
	\label{alg_Algorithm2}
\end{algorithm}

The gated block is adopted to focus on useful information while filtering out redundant information. This process can be expressed as
\begin{equation}
	{\bf Z} = {\hat {\bf Z}_1^{0}} \odot \sigma_2({\bf Z}_2)
\end{equation}
where $\odot$ denotes element-wise multiplication. Finally, an FC layer is used to adjust the token size, and a residual connection is applied to reduce the loss of information:
\begin{equation}
	{\bf S}_i = {\bf S}_{i-1} + {\rm Linear}_{N}({\bf Z}).
\end{equation}

An overview of the proposed pyramid SSM is summarized in Algorithm \ref{alg_Algorithm1} and Algorithm \ref{alg_Algorithm2}, where $P$ denotes the batch size, $\otimes$ denotes the outer product, and ${\rm Parameter}^{\bf o}$ denotes the learnable parameter.
\subsection{Spectral Contrastive Head}
The SCH further projects the extracted multiresolution spectral features into the feature space using two FC layers, which can be expressed as 
\begin{equation}
	{\bf x''} = {\rm Linear}_{d}(\sigma_1({\rm Linear}_{2d}({\bf x'})))
\end{equation}
where ${\bf x'} \in \mathbb{R}^{LC}$ denotes the reshaped spectral features, and $d$ denotes the size of feature vectors.

To maximize the similarity of positive samples while minimizing the similarity of negative samples, contrastive learning is required. In this paper, the feature vector pairs $({\bf x''}, {\bf y''})$ from the same location construct the positive samples, and those from different locations construct the negative ones. For example, given a batch of  $({\bf x''}, {\bf y''})$,  for a specific sample ${\bf x''}^{(i)}$, there are $P$ pairs in total, where $({\bf x''}^{(i)}, {\bf y''}^{(i)})$ is the only positive pair while the others are negative pairs. Thus, the loss for ${\bf x''}^{(i)}$ can be expressed as 
\begin{equation}
	\ell({\bf x''}^{(i)},  {\bf y''}^{(i)}) = -{\rm log}\frac{{\rm exp}(c({\bf x''}^{(i)},  {\bf y''}^{(i)})/\alpha)}{ \sum_{j=1}^P{{\rm exp}(c({\bf x''}^{(i)},  {\bf y''}^{(j)})/\alpha)}}
\end{equation}
where $\alpha$ is the temperature parameter to control the degree of attention to negative pairs. Taking into account learning the similarity of all the positive pairs within a batch, the spectral contrastive loss can be expressed as 
\begin{equation}
	L_{\rm contrast} = \frac{1}{P} \sum_{i=1}^P\ell({\bf x''}^{(i)},  {\bf y''}^{(i)}).
\end{equation}

Through spectrally contrastive learning, similar samples are brought closer together while dissimilar samples are pushed further apart, enabling the model to distinguish the spectral similarities and dissimilarities of different instances.
\section{Experimental Results and Analysis}
\label{experiments}
In this section, we compare the proposed HTD-Mamba method with state-of-the-art methods on four benchmark datasets to verify its effectiveness.
\subsection{Datasets and Experimental Setup}
\subsubsection{Datasets} The first two datasets, San Diego I and II, were captured by the Airborne Visible/Infrared Imaging Spectrometer (AVIRIS) over San Diego Airport, San Diego, California, USA. Both HSIs have a size of $100 \times 100 \times 224$ with a spatial resolution of 3.5 m, covering a wavelength range from 400 to 2,500 nm. After removing low-quality bands (1–6, 33–35, 97, 107–113, 153–166, and 221–224), 189 bands are preserved for experiments. The targets to be detected are three airplanes, consisting of 58 pixels in San Diego I and 134 pixels in San Diego II, respectively.

The third dataset, Los Angeles, was collected by AVIRIS over an airport scene in Los Angeles, California, USA. The HSI has a size of $100\times100\times205$, with $100\times100$ pixels and 205 spectral bands, covering a wavelength range from 400 to 2,500 nm. The spatial resolution of this dataset is 7.1 m. The targets to be detected are two airplanes, consisting of 87 pixels.

The fourth dataset, Pavia, was captured by the Reflective Optics System Imaging Spectrometer (ROSIS-03) sensor over a beach area in Pavia, Italy. The HSI has a spatial size of $150\times150$ with a spatial resolution of 1.3 m. It consists of 115 spectral bands ranging from 430 nm to 860 nm. After removing low-quality bands, 102 bands remain for the experiments. Some man-made objects are labeled as targets, consisting of 68 pixels. Both Los Angeles and Pavia datasets are available online\footnote[1]{\url{http://xudongkang.weebly.com/}}.

In the experiments, the pixel closest to the averaged spectrum of all the target pixels is selected as the target spectrum. The false-color images and the ground-truth maps are displayed in Fig. \ref{detection_results}.
\subsubsection{Quality Metrics}
To assess the quality of the detection map, both qualitative and quantitative metrics are adopted. Specifically, the visualized detection map, receiver operating characteristic (ROC) curve, and background-target separability diagram are utilized for qualitative assessment. The ROC curve is a function of false alarm rate $P_f$ and detection probability $P_d$ under different thresholds $\tau$. For comprehensive analysis, 3D ROC curves \cite{chang2020effective} are adopted, where three unfolded 2D ROC curves of $(P_f, P_d)$, $(P_f, \tau)$, and $(P_d, \tau)$ are obtained to measure detection effectiveness, target detectability, and background suppression, respectively. The background-target separability diagram is plotted using the Box–Whisker function to show the distinction between background and target. For quantitative evaluation, three areas under the ROC curves (AUCs) are applied, i.e., ${\rm AUC}(P_f, P_d)$, ${\rm AUC}(\tau, P_d)$, and ${\rm AUC}(\tau, P_f)$. In addition, two composite metrics are also calculated to evaluate overall quality:
\begin{equation}
	{\rm AUC_{OA}} = {\rm AUC}(P_f, P_d) + {\rm AUC}(\tau, P_d) - {\rm AUC}(\tau, P_f),
\end{equation}
\begin{equation}
	{\rm AUC_{SNPR}} = {\rm AUC}(\tau, P_d) / {\rm AUC}(\tau, P_f).
\end{equation}
The detailed properties of these five metrics are listed in Table \ref{metrics}.
\begin{table}[!t]
	\centering
	\caption{Properties of quantitative metrics}
	\scriptsize
	\tabcolsep = 1.3pt
	\renewcommand\arraystretch{1.5}
	\begin{tabular}{c|ccccc}
		\hline
		Metric&$\rm {AUC}{(P_f, P_d)}$&$\rm {AUC}{(\tau, P_d)}$&$\rm {AUC}{(\tau, P_f)}$&$\rm {AUC}_{OA}$&$\rm {AUC}_{SNPR}$\\ 
		\hline
		Perspective&Effectiveness&Detectability&False alarm&Overall&Overall\\
		Range&[0, 1]&[0, 1]&[0, 1]&[-1, 2]&[0, +$\infty$)\\
		Trend&$\uparrow$&$\uparrow$&$\downarrow$&$\uparrow$&$\uparrow$\\
		Optimum&1&1&0&2&+$\infty$\\
		\hline
	\end{tabular}
	\label{metrics}
\end{table}
\subsubsection{Implementation Details}
The proposed HTD-Mamba is implemented using Python 3.10.4 and PyTorch 2.0.1, with an Intel Core i9-10900X CPU, 32 GB of RAM, and an NVIDIA TITAN RTX with 24 GB of GPU memory. The batch size $P$ is set to 80. For each batch, the patch size surrounding the center pixel is set to $11\times11$, $13\times13$, $11\times11$, and $5\times5$ for San Diego I, San Diego II, Los Angeles, and Pavia, respectively. In the group-wise spectral embedding, the group length $m$ is set to 30, 5, 5, and 15 for the four datasets, respectively. The embedding size $N$ is set to 16 for all four datasets. In the feature extraction and projection blocks, the depth of the pyramid SSM is set to 1, the hidden state size $D$ is set to 16, and the dimension of the final feature vectors $d$ is set to 32 for all four datasets. For the contrastive loss, the temperature parameter $\alpha$ is fixed at 0.1 for all four datasets. The AdamW optimizer is selected for training, where the weight-decay parameter is set to $10^{-4}$. The initial learning rate is set to $10^{-4}$, which increases via a linear warmup scheduler in the first $10\%$ epochs and then decreases via a cosine scheduler. The total number of epochs is 200. In the background suppression process, the parameter $\delta$ is set to 0.1 for all four datasets.
\begin{figure*}[!t]
	\centering
	\footnotesize
	\vspace{-3pt}
	\subfigure{
		\begin{minipage}[c]{0.05\textwidth}
			\centering
			\text{(a)}
		\end{minipage}
		\hspace{-5pt}
		\begin{minipage}[c]{1.0\textwidth}
			\includegraphics[width=0.08\textwidth]{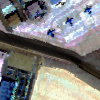}
			\includegraphics[width=0.08\textwidth]{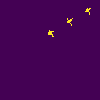} 
			\includegraphics[width=0.08\textwidth]{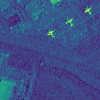} 
			\includegraphics[width=0.08\textwidth]{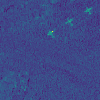} 
			\includegraphics[width=0.08\textwidth]{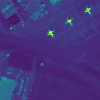} 
			\includegraphics[width=0.08\textwidth]{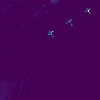}
			\includegraphics[width=0.08\textwidth]{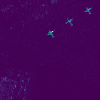} 
			\includegraphics[width=0.08\textwidth]{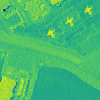} 
			\includegraphics[width=0.08\textwidth]{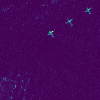} 
			\includegraphics[width=0.08\textwidth]{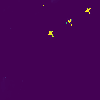} 
			\includegraphics[width=0.08\textwidth]{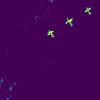} 
		\end{minipage}
	}\\
	\vspace{-3pt}
	\subfigure{
		\begin{minipage}[c]{0.05\textwidth}
			\centering
			\text{(b)}
		\end{minipage}
		\hspace{-5pt}
		\begin{minipage}[c]{1.0\textwidth}
			\includegraphics[width=0.08\textwidth]{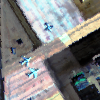}
			\includegraphics[width=0.08\textwidth]{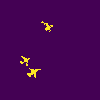} 
			\includegraphics[width=0.08\textwidth]{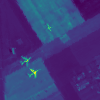} 
			\includegraphics[width=0.08\textwidth]{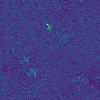} 
			\includegraphics[width=0.08\textwidth]{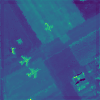} 
			\includegraphics[width=0.08\textwidth]{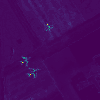}
			\includegraphics[width=0.08\textwidth]{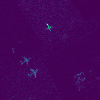} 
			\includegraphics[width=0.08\textwidth]{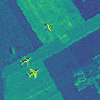} 
			\includegraphics[width=0.08\textwidth]{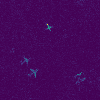} 
			\includegraphics[width=0.08\textwidth]{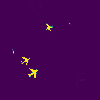} 
			\includegraphics[width=0.08\textwidth]{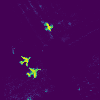} 
		\end{minipage}
	}\\
	\vspace{-3pt}
	\subfigure{
		\begin{minipage}[c]{0.05\textwidth}
			\centering
			\text{(c)}
		\end{minipage}
		\hspace{-5pt}
		\begin{minipage}[c]{1.0\textwidth}
			\includegraphics[width=0.08\textwidth]{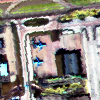}
			\includegraphics[width=0.08\textwidth]{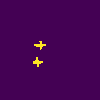} 
			\includegraphics[width=0.08\textwidth]{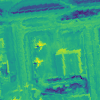} 
			\includegraphics[width=0.08\textwidth]{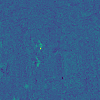} 
			\includegraphics[width=0.08\textwidth]{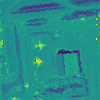} 
			\includegraphics[width=0.08\textwidth]{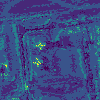}
			\includegraphics[width=0.08\textwidth]{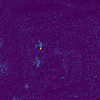} 
			\includegraphics[width=0.08\textwidth]{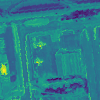} 
			\includegraphics[width=0.08\textwidth]{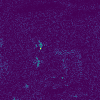} 
			\includegraphics[width=0.08\textwidth]{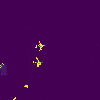} 
			\includegraphics[width=0.08\textwidth]{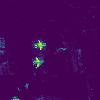}
		\end{minipage}
	}\\
	\vspace{-3pt}
	\subfigure{
		\begin{minipage}[c]{0.05\textwidth}
			\centering
			\text{(d)}
		\end{minipage}
		\hspace{-5pt}
		\begin{minipage}[c]{1.0\textwidth}
			\includegraphics[width=0.08\textwidth]{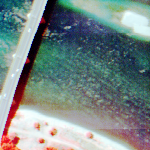}
			\includegraphics[width=0.08\textwidth]{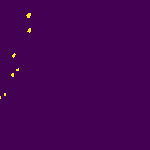} 
			\includegraphics[width=0.08\textwidth]{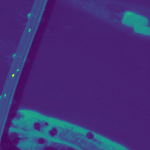} 
			\includegraphics[width=0.08\textwidth]{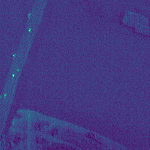} 
			\includegraphics[width=0.08\textwidth]{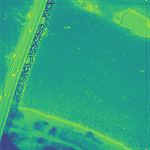} 
			\includegraphics[width=0.08\textwidth]{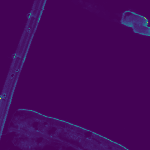}
			\includegraphics[width=0.08\textwidth]{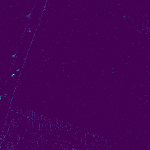} 
			\includegraphics[width=0.08\textwidth]{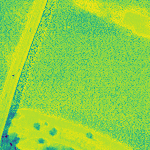} 
			\includegraphics[width=0.08\textwidth]{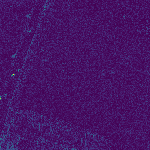} 
			\includegraphics[width=0.08\textwidth]{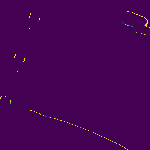} 
			\includegraphics[width=0.08\textwidth]{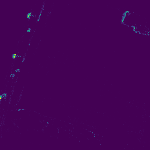}
		\end{minipage}
	}\\
	\vspace{-3pt}
	\subfigure{
		\begin{minipage}[c]{0.05\textwidth}
			\centering
			\text{}
		\end{minipage}
		\hspace{-5pt}
		\begin{minipage}[c]{1.0\textwidth}
			\parbox{0.08\textwidth}{\centering \text{False-color}}
			\parbox{0.08\textwidth}{\centering \text{Ground truth}}
			\parbox{0.08\textwidth}{\centering \text{OSP}}
			\parbox{0.08\textwidth}{\centering \text{CEM}}
			\parbox{0.08\textwidth}{\centering \text{CSCR}}
			\parbox{0.08\textwidth}{\centering \text{CTTD}}
			\parbox{0.08\textwidth}{\centering \text{BLTSC}}
			\parbox{0.08\textwidth}{\centering \text{MLSN}}
			\parbox{0.08\textwidth}{\centering \text{OS-VAE}}
			\parbox{0.08\textwidth}{\centering \text{TSTTD}}
			\parbox{0.08\textwidth}{\centering \text{HTD-Mamba}}
		\end{minipage}
	} \caption{Visualized results of the competing methods on (a) San Diego I, (b) San Diego II, (c) Los Angeles, and (d) Pavia.}
	\label{detection_results}
\end{figure*}
\begin{figure*}[!t]
	\centering
	\subfigure{
		\includegraphics[width=0.88\textwidth]{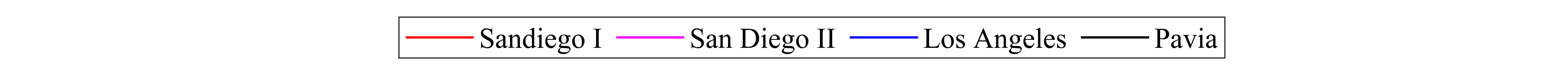}
	}
	\renewcommand{\thesubfigure}{(a)}
	\subfigure[]{
		\includegraphics[width=0.22\textwidth]{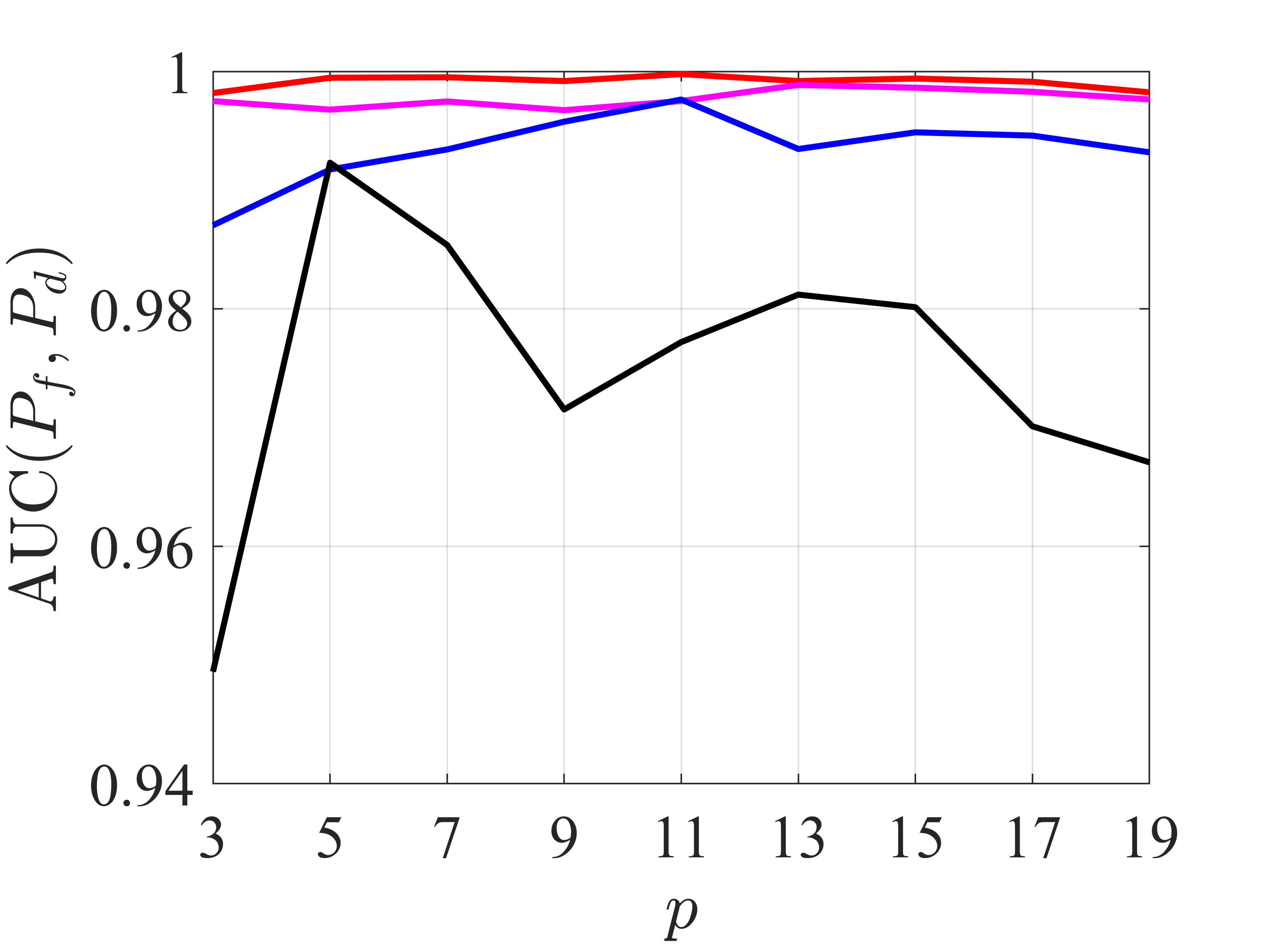}
	}
	\renewcommand{\thesubfigure}{(b)}
	\subfigure[]{
		\includegraphics[width=0.22\textwidth]{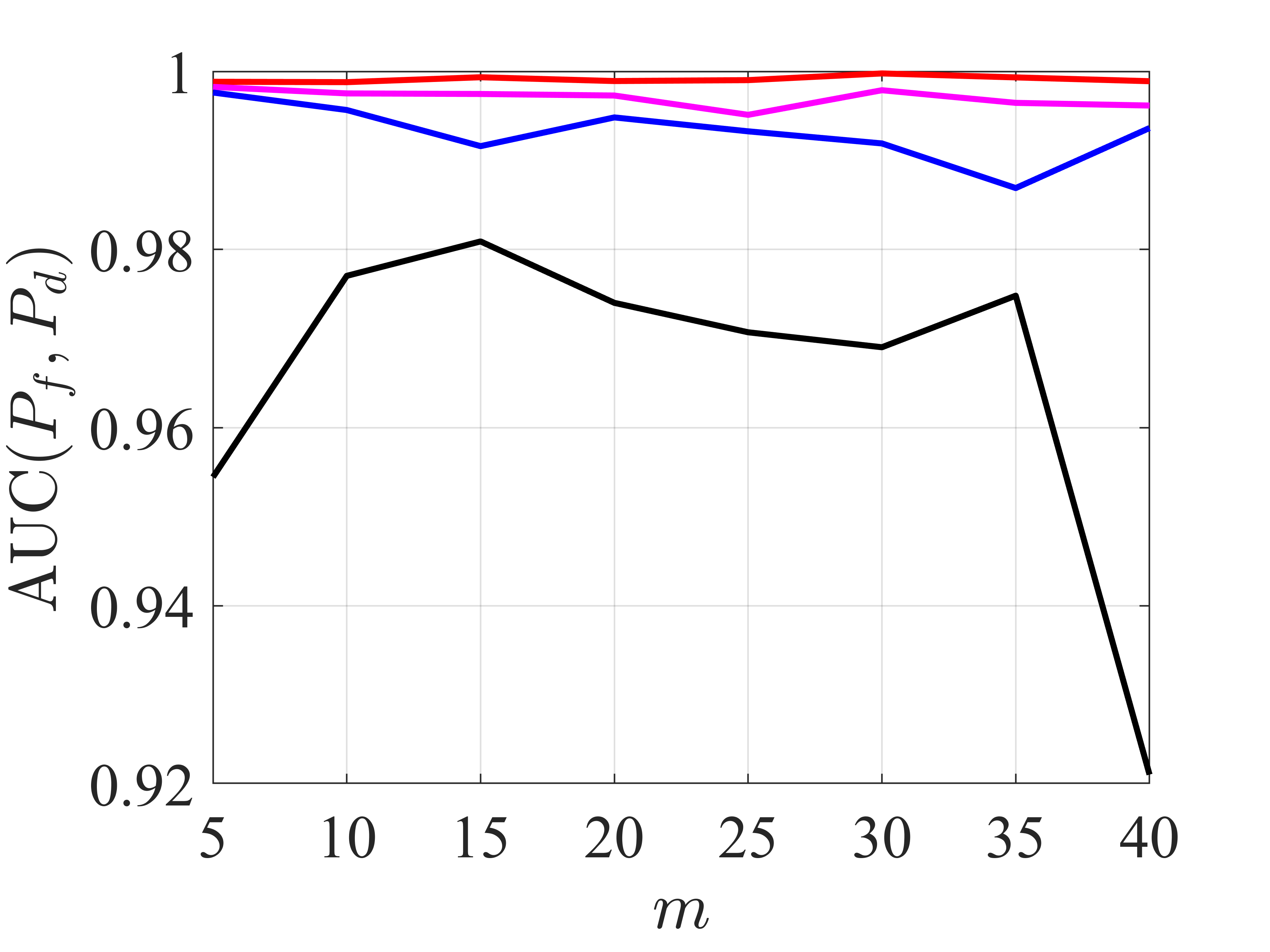}
	}
	\renewcommand{\thesubfigure}{(c)}
	\subfigure[]{
		\includegraphics[width=0.22\textwidth]{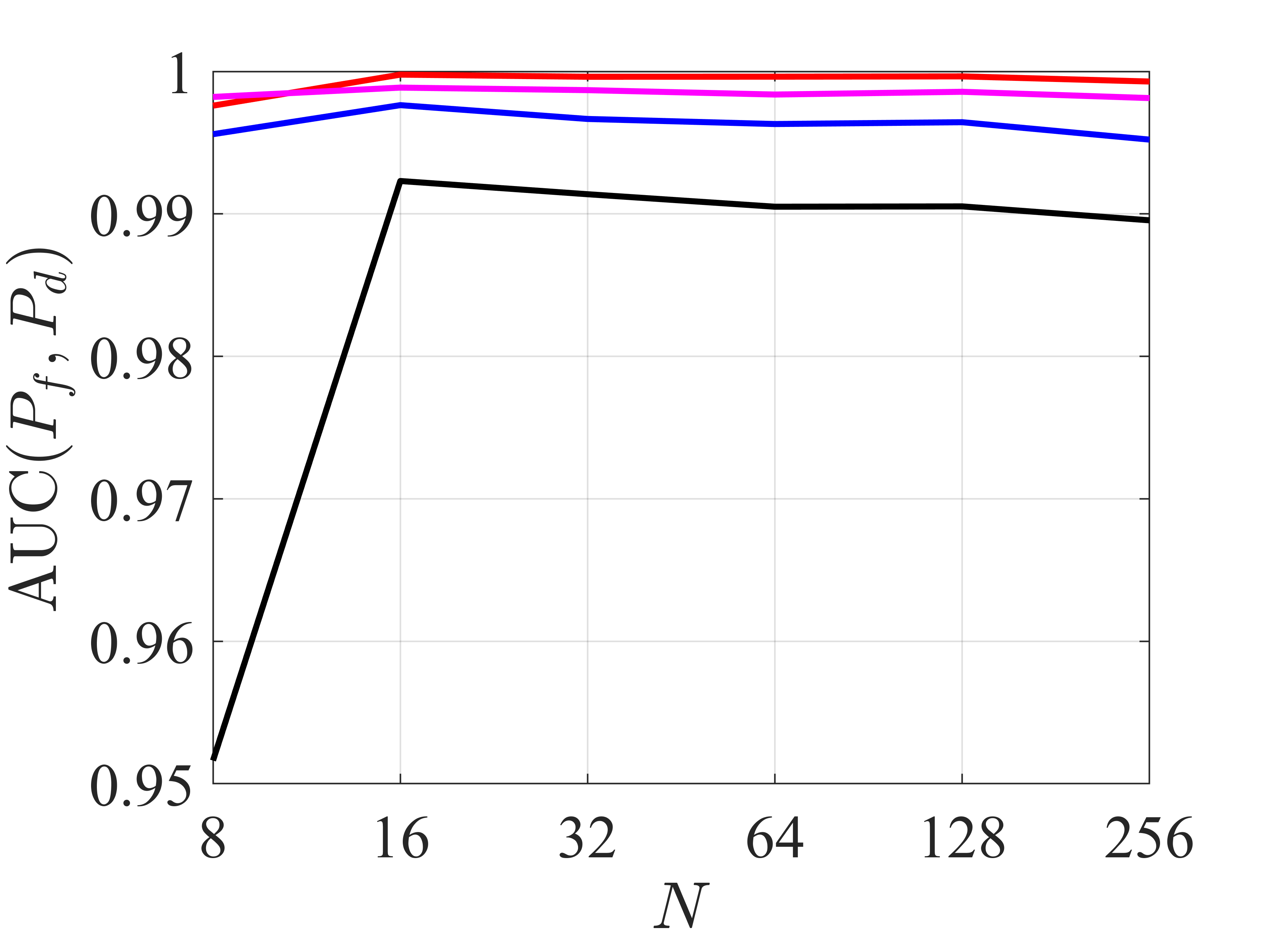}
	}
	\renewcommand{\thesubfigure}{(d)}
	\subfigure[]{
		\includegraphics[width=0.22\textwidth]{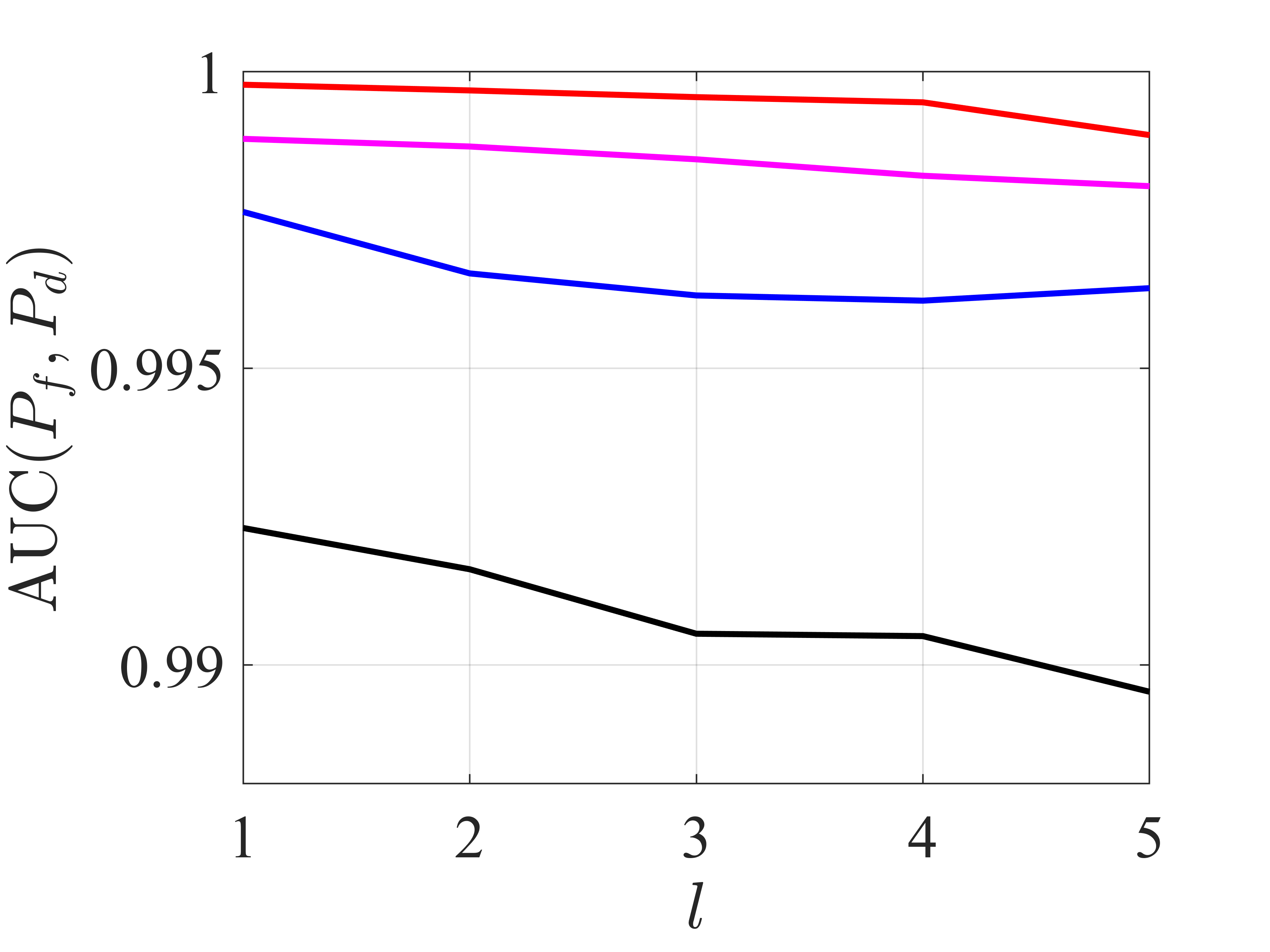}
	}
	\renewcommand{\thesubfigure}{(\alph{subfigure})}
	\caption{Parameter effect analysis of (a) patch size $p$, (b) spectral group length $m$, (c) embedding size $N$, and (d) network depth $l$.}
	\label{parameter_analysis}
\end{figure*}
\begin{figure*}[htbp]
	\centering
	\begin{minipage}{1.0\textwidth}
		\subfigure{
			\includegraphics[width=0.25\textwidth]{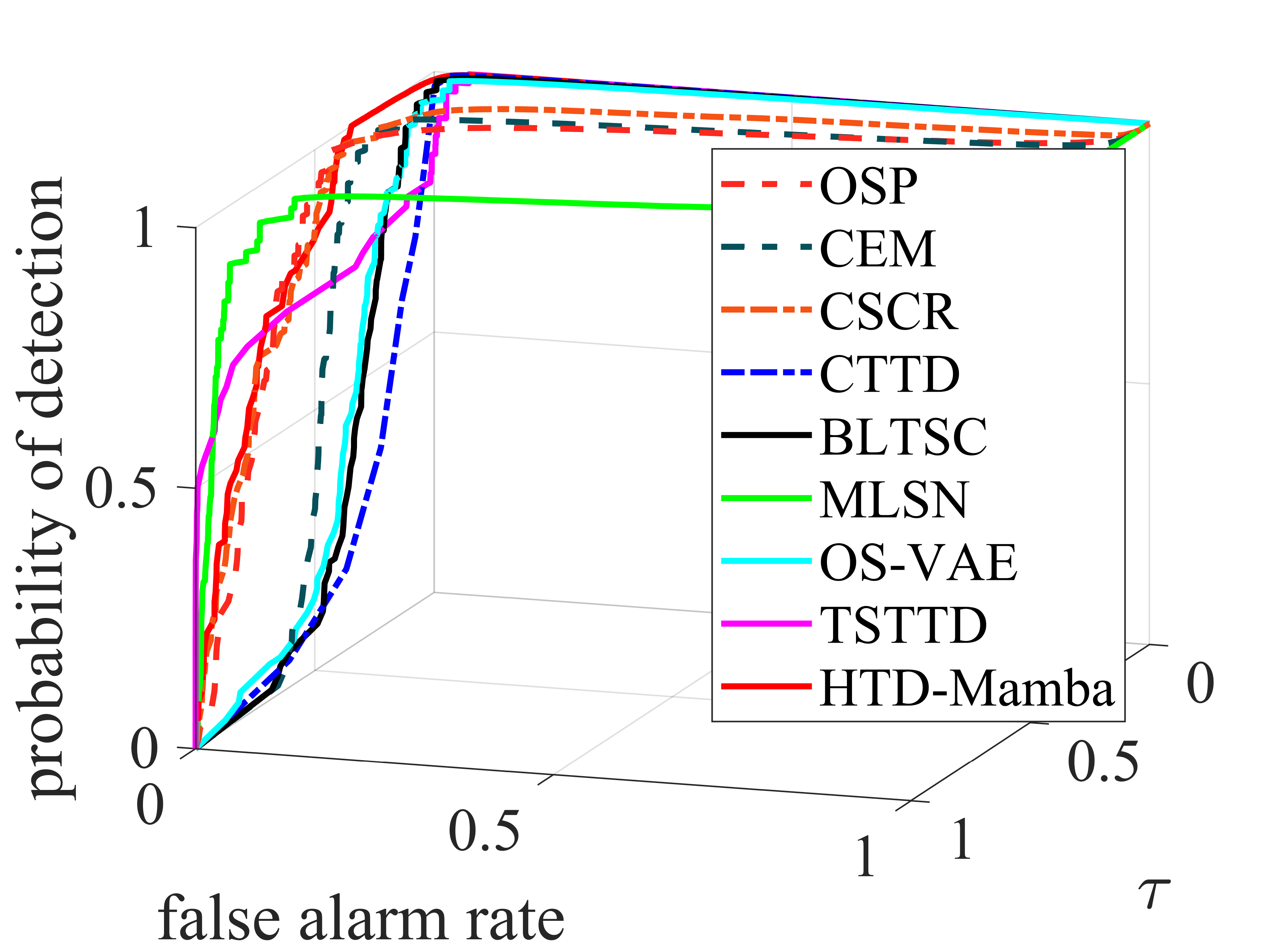}
		}
		\hspace{-18pt}
		\subfigure{
			\includegraphics[width=0.25\textwidth]{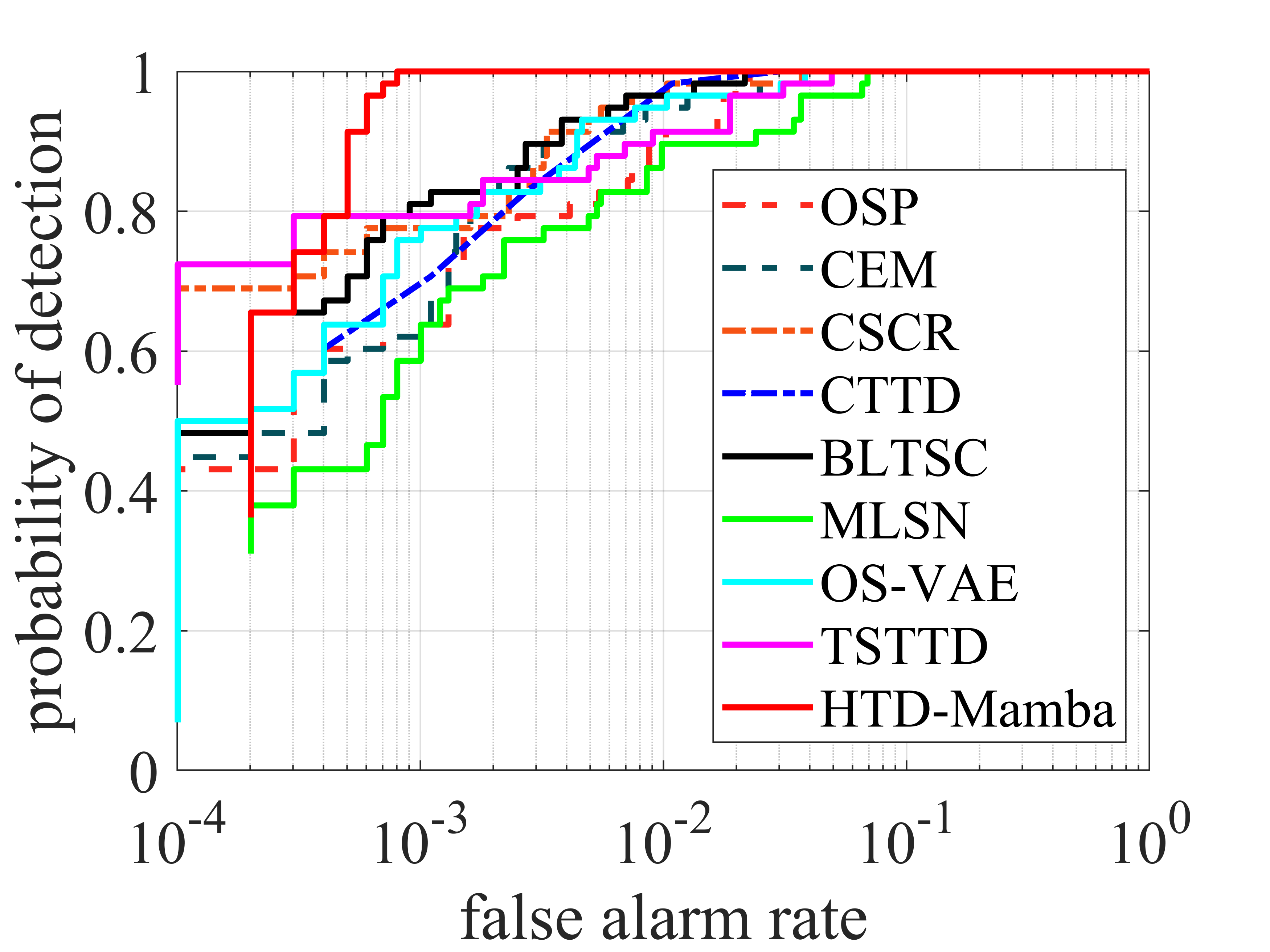}
		}
		\hspace{-18pt}
		\subfigure{
			\includegraphics[width=0.25\textwidth]{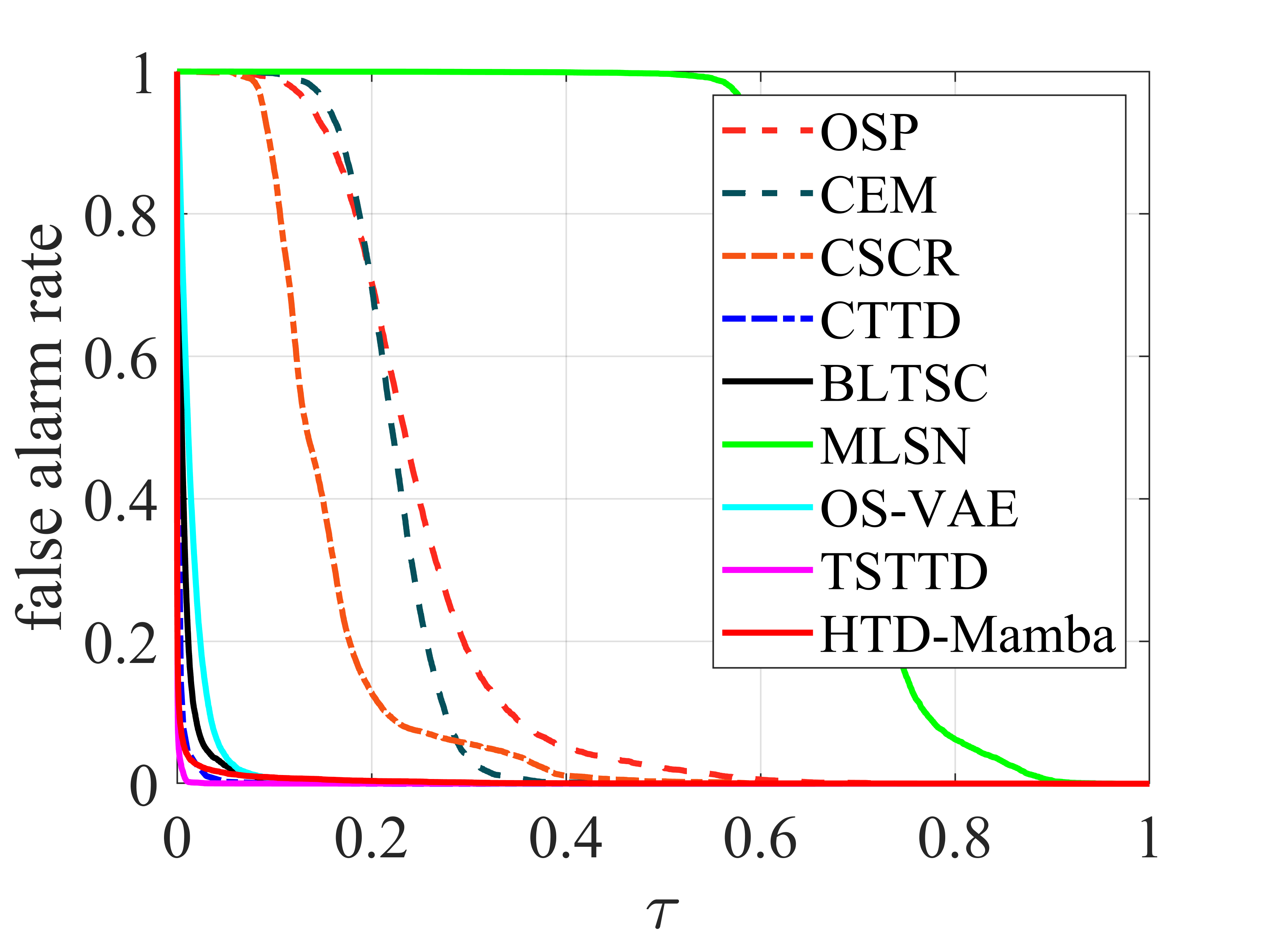}
		}
		\hspace{-18pt}
		\subfigure{
			\includegraphics[width=0.25\textwidth]{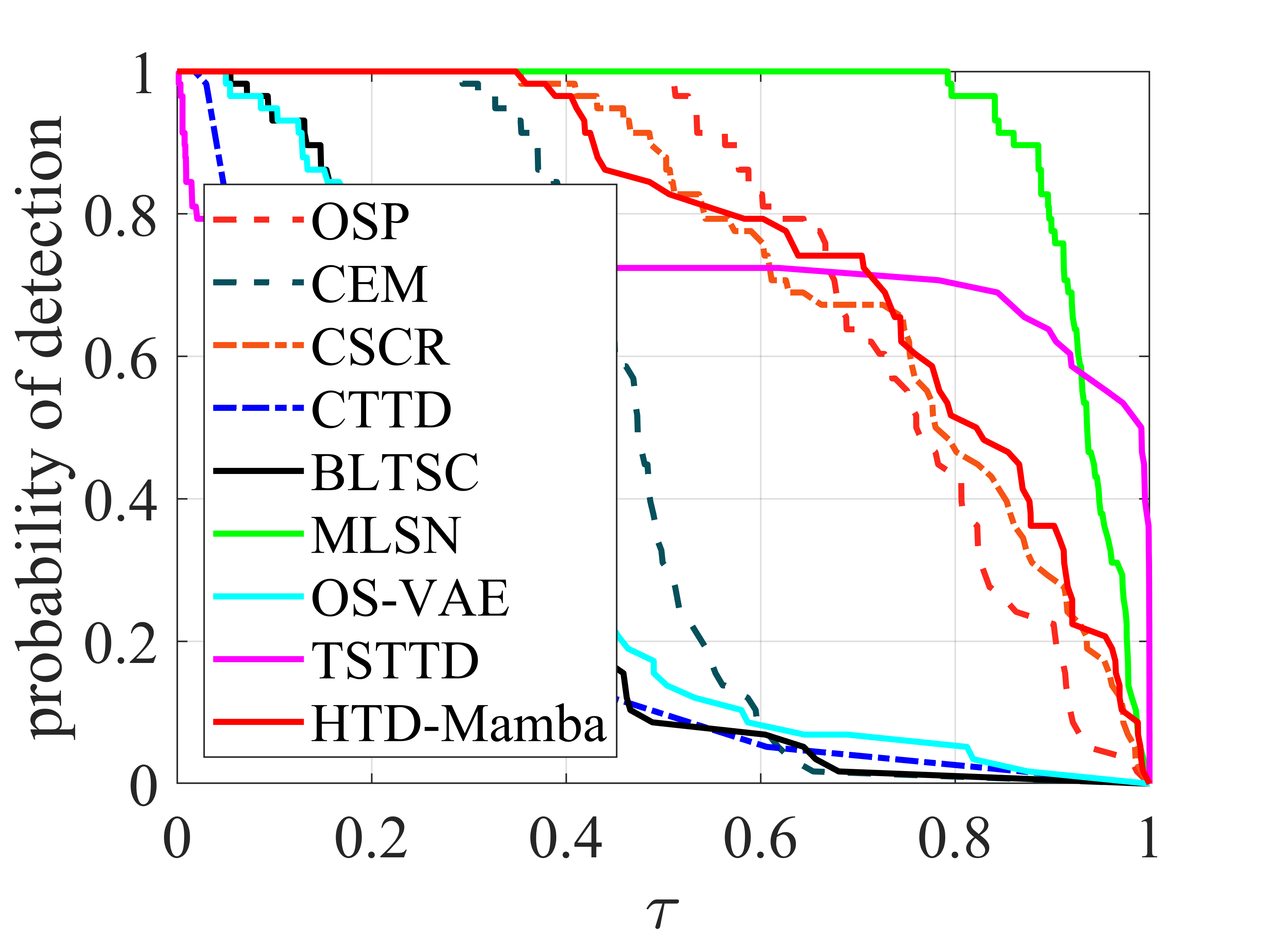}
		}\\
		\parbox{1.0\textwidth}{\centering {\footnotesize (a)}}
	\end{minipage}
	\begin{minipage}{1.0\textwidth}
		\subfigure{
			\includegraphics[width=0.25\textwidth]{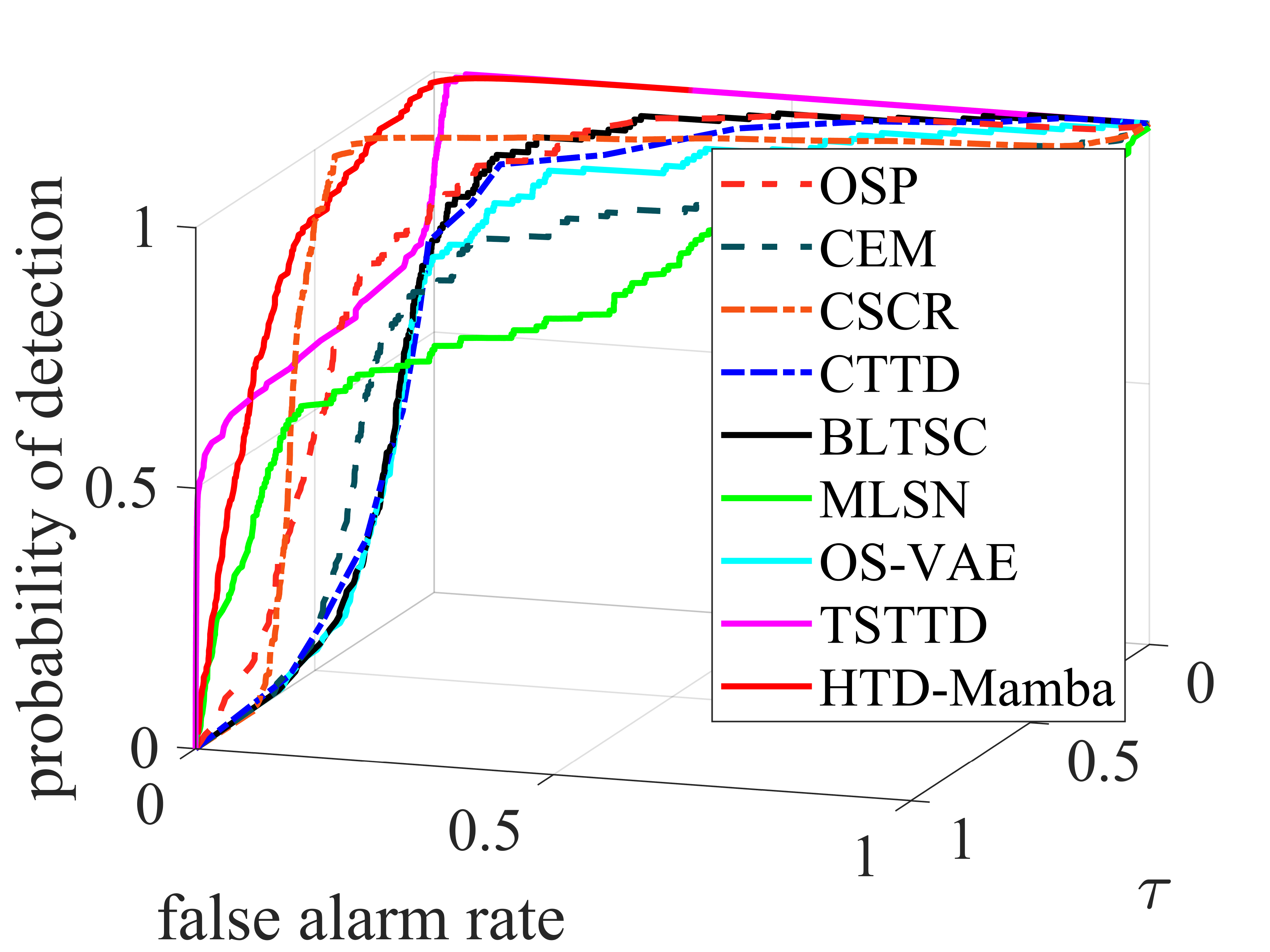}
		}
		\hspace{-18pt}
		\subfigure{
			\includegraphics[width=0.25\textwidth]{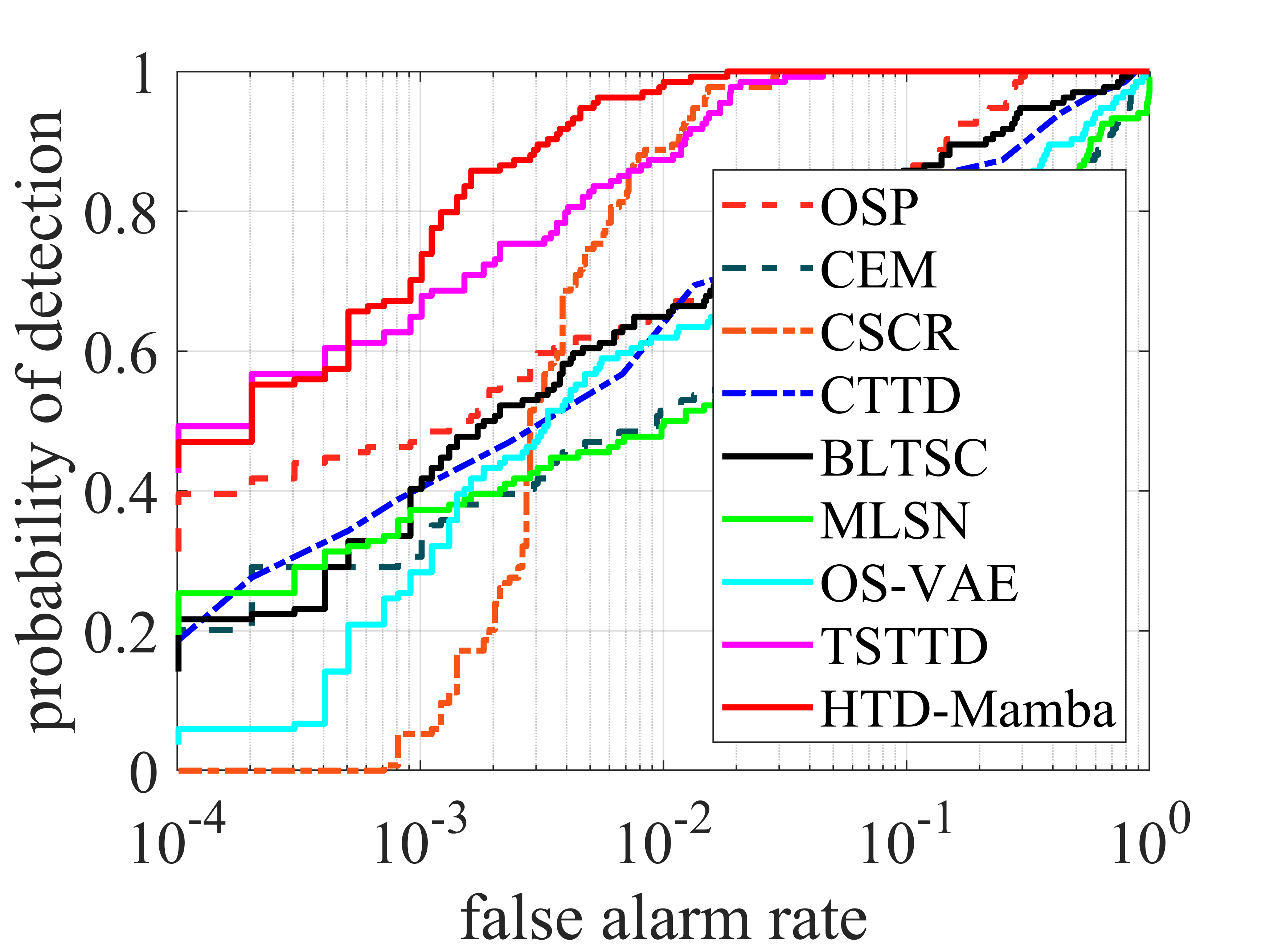}
		}
		\hspace{-18pt}
		\subfigure{
			\includegraphics[width=0.25\textwidth]{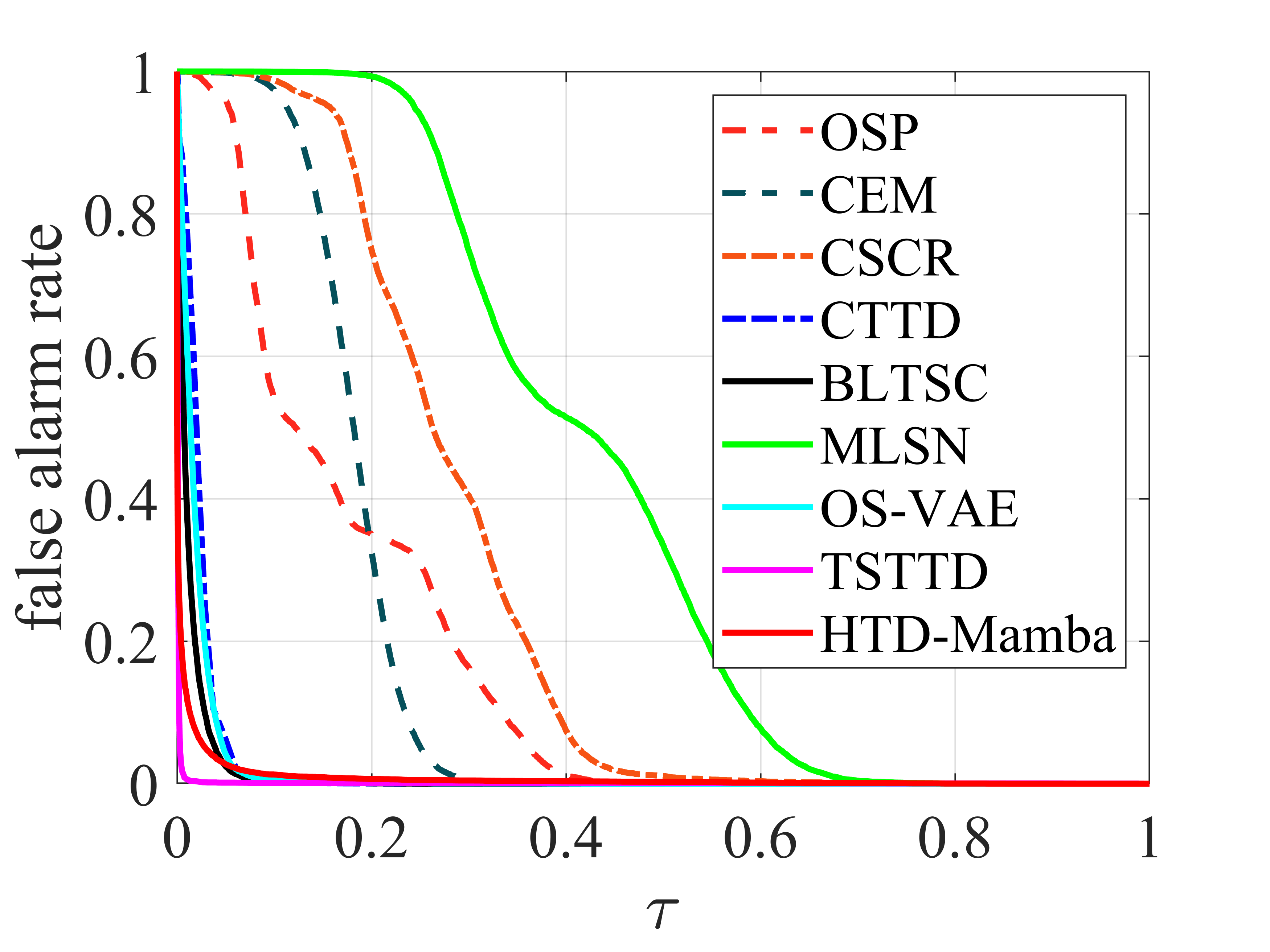}
		}
		\hspace{-18pt}
		\subfigure{
			\includegraphics[width=0.25\textwidth]{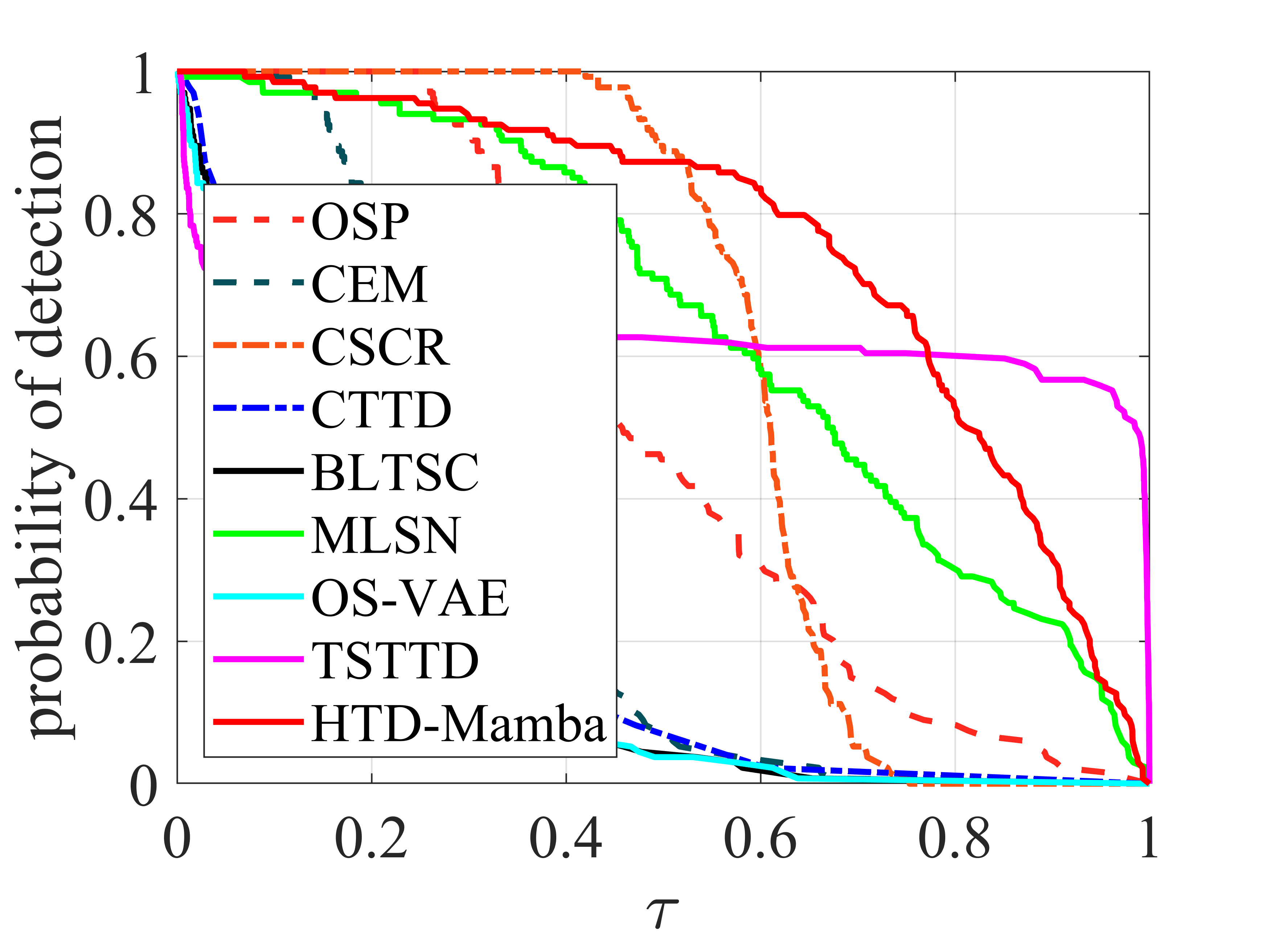}
		}\\
		\parbox{1.0\textwidth}{\centering {\footnotesize (b)}}
	\end{minipage}
	\begin{minipage}{1.0\textwidth}
		\subfigure{
			\includegraphics[width=0.25\textwidth]{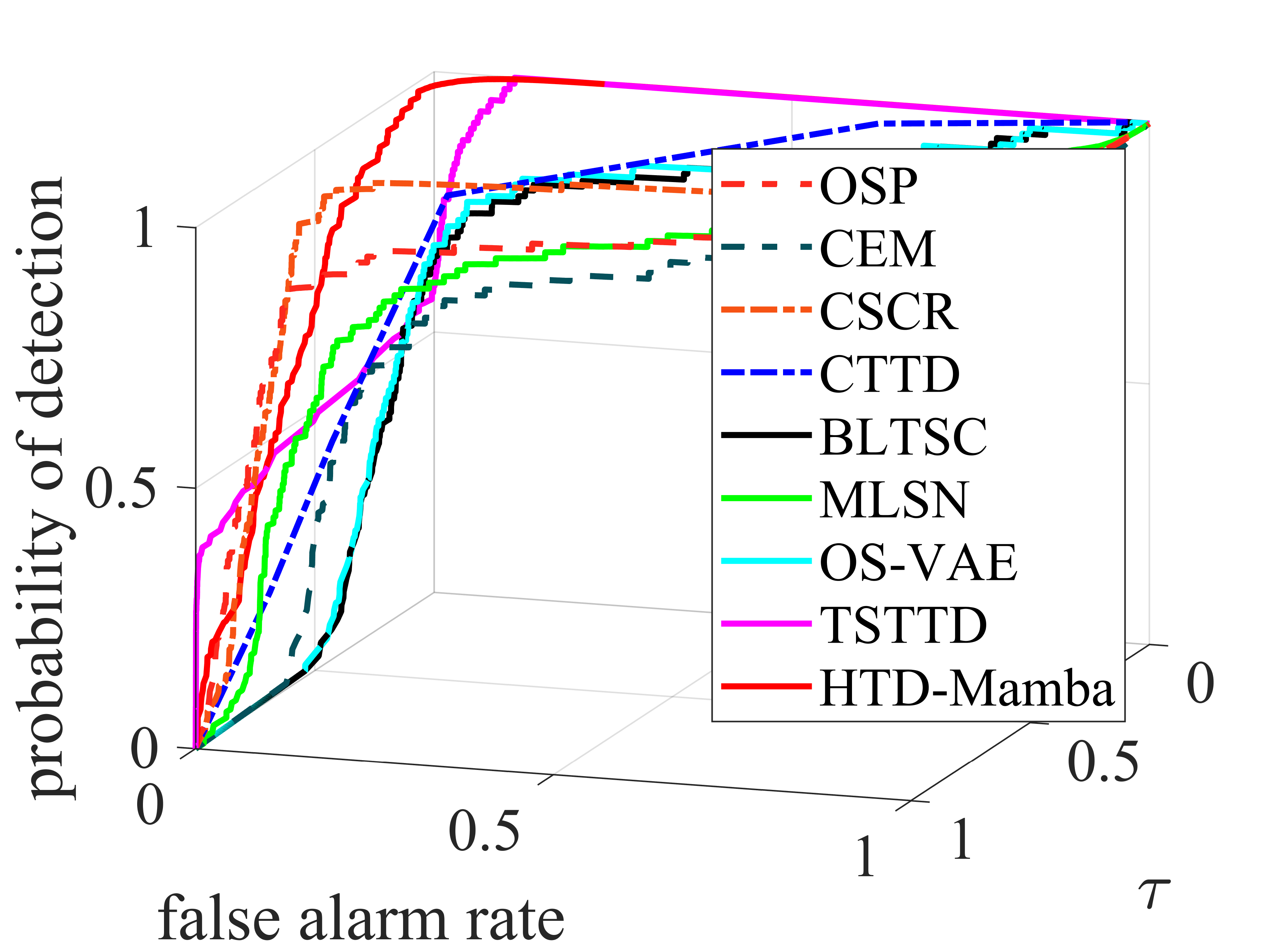}
		}
		\hspace{-18pt}
		\subfigure{
			\includegraphics[width=0.25\textwidth]{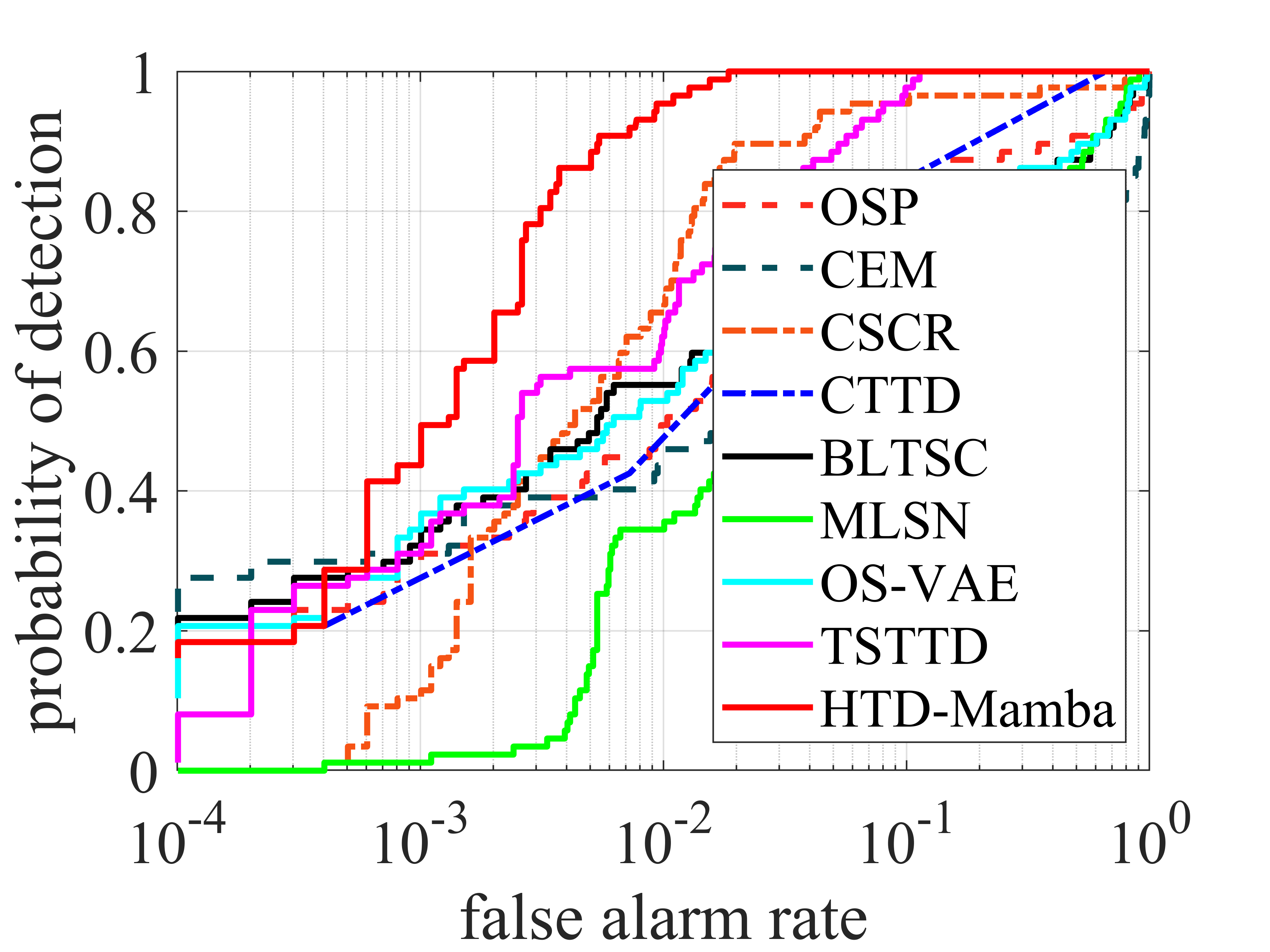}
		}
		\hspace{-18pt}
		\subfigure{
			\includegraphics[width=0.25\textwidth]{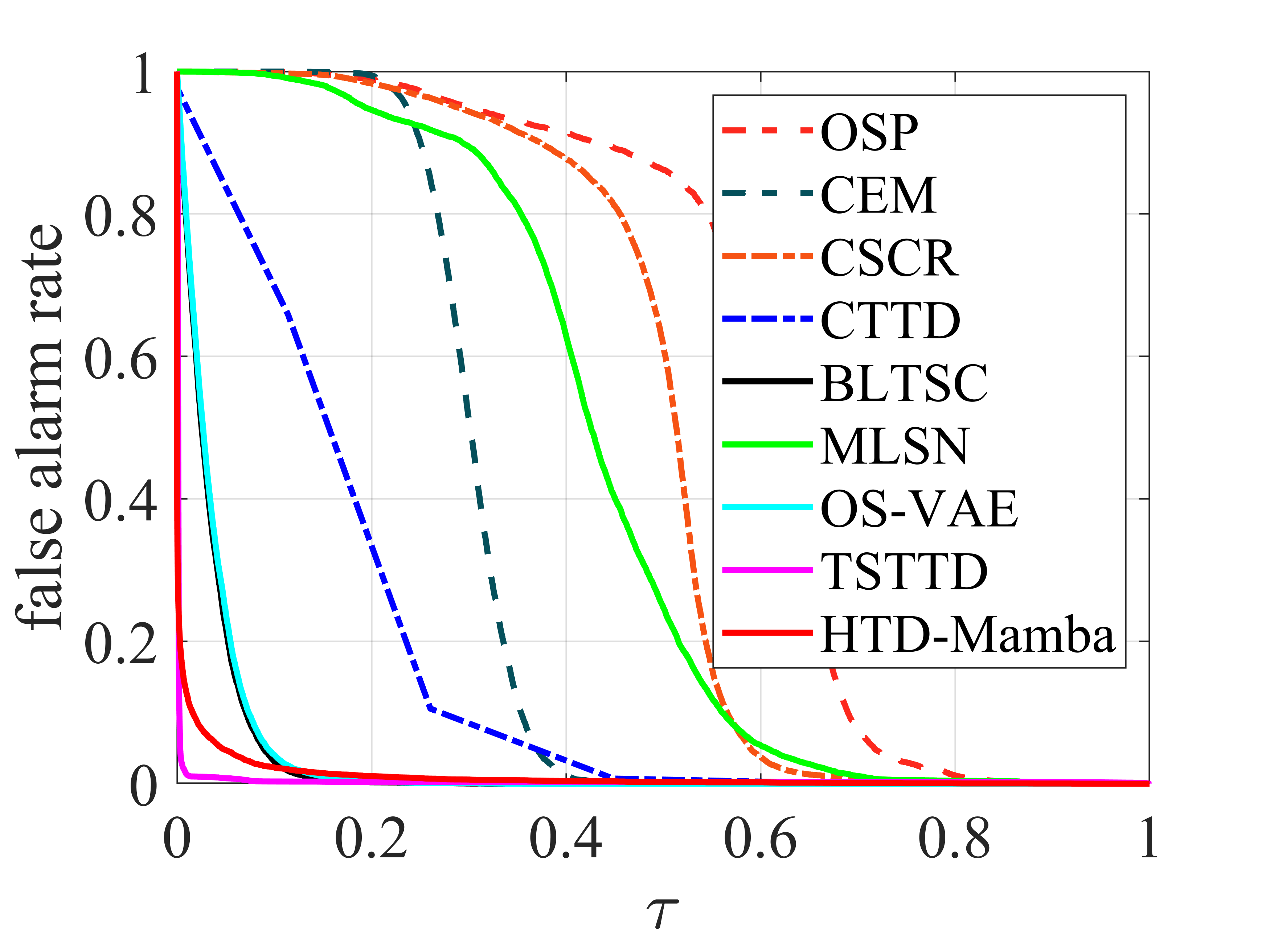}
		}
		\hspace{-18pt}
		\subfigure{
			\includegraphics[width=0.25\textwidth]{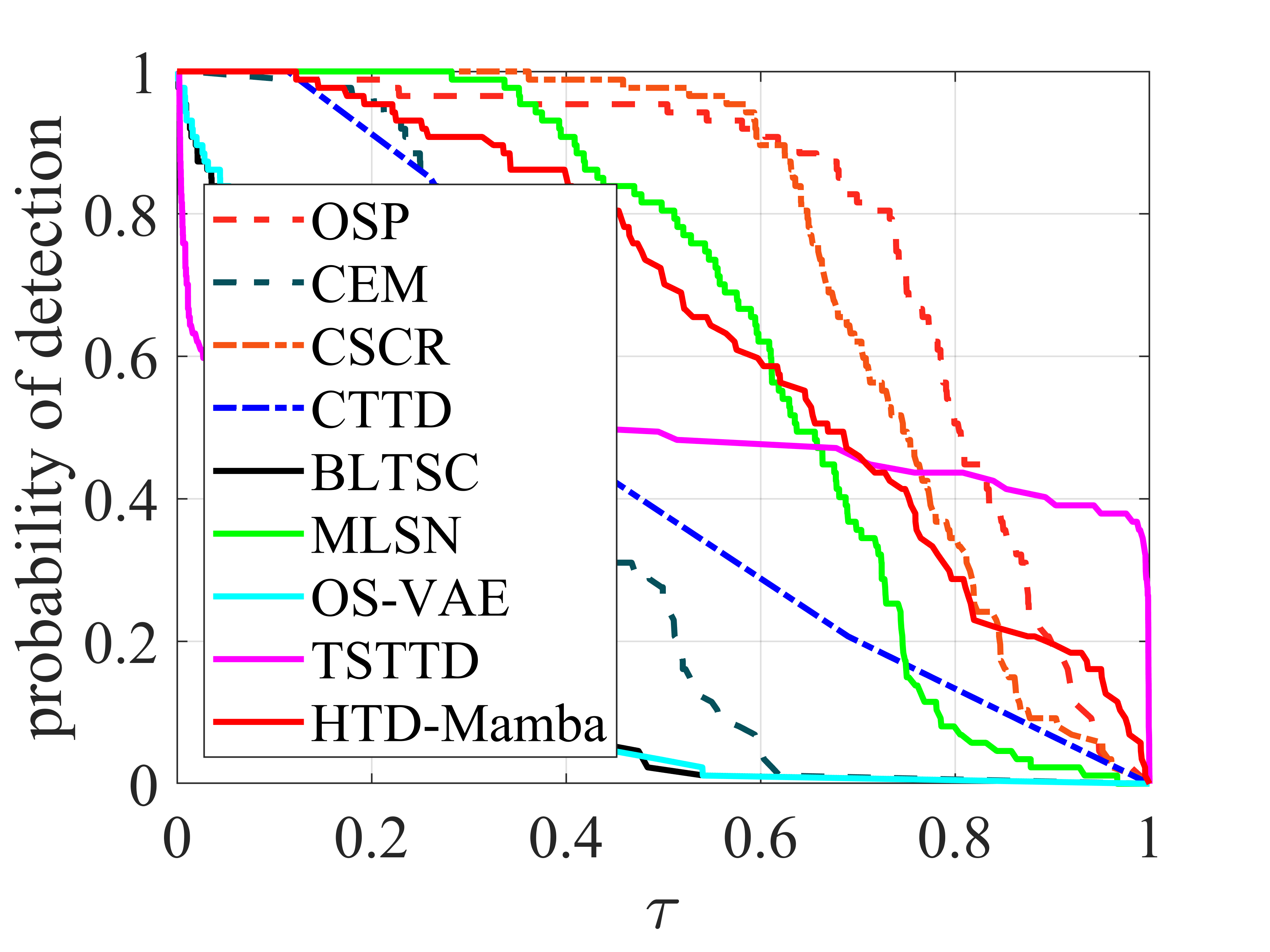}
		}\\
		\parbox{1.0\textwidth}{\centering {\footnotesize (c)}}
	\end{minipage}
	\begin{minipage}{1.0\textwidth}
		\subfigure{
			\includegraphics[width=0.25\textwidth]{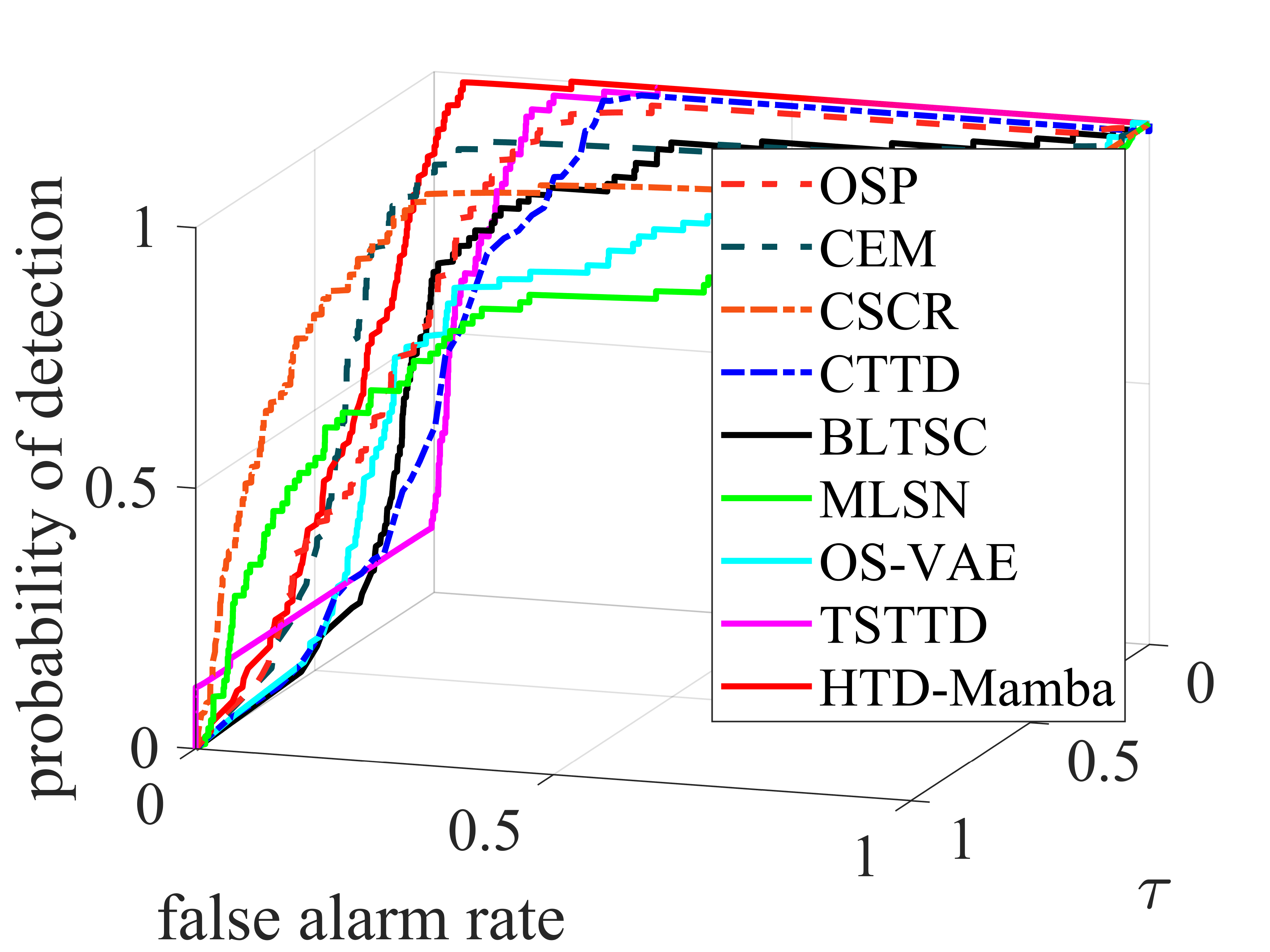}
		}
		\hspace{-18pt}
		\subfigure{
			\includegraphics[width=0.25\textwidth]{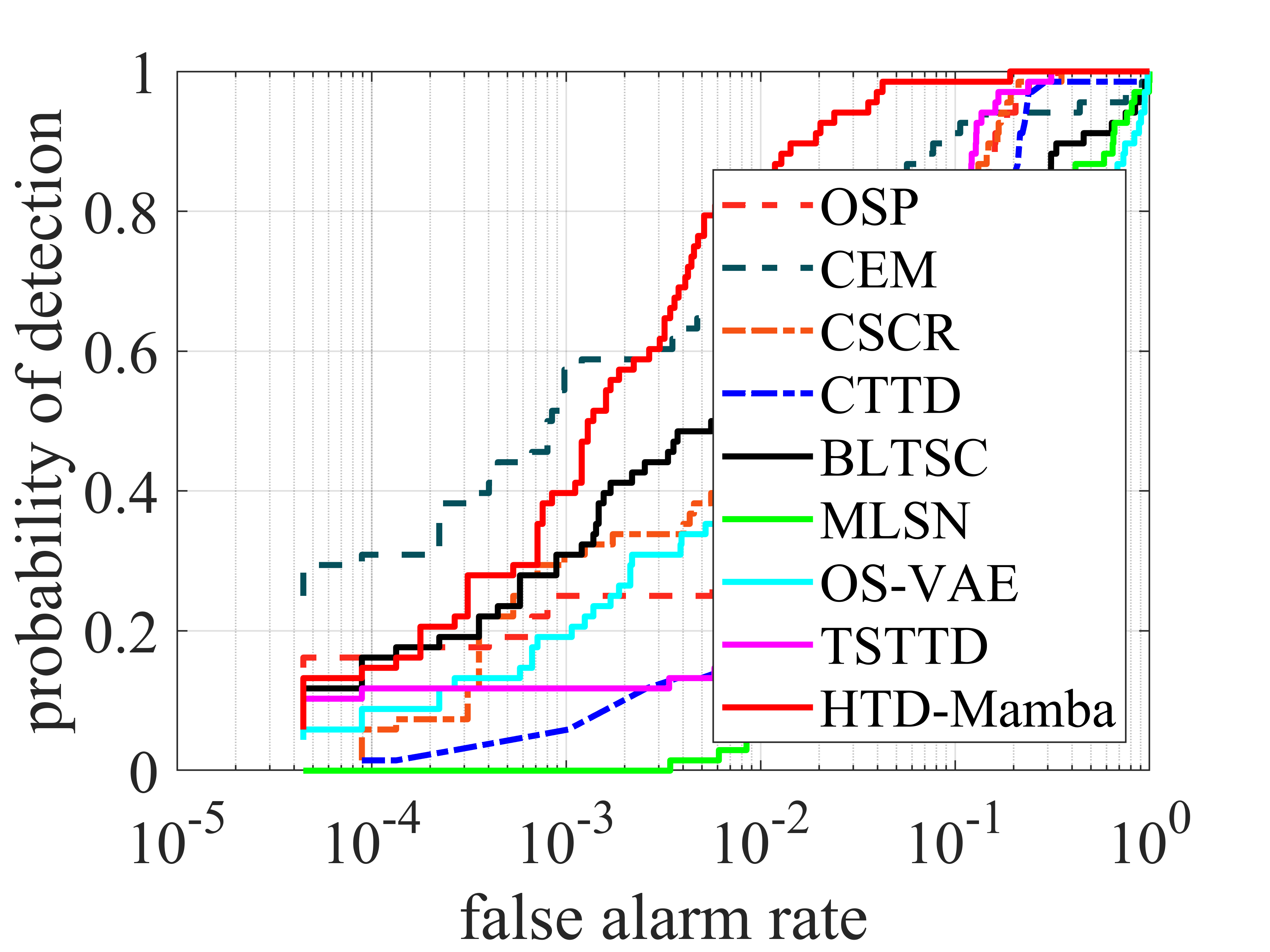}
		}
		\hspace{-18pt}
		\subfigure{
			\includegraphics[width=0.25\textwidth]{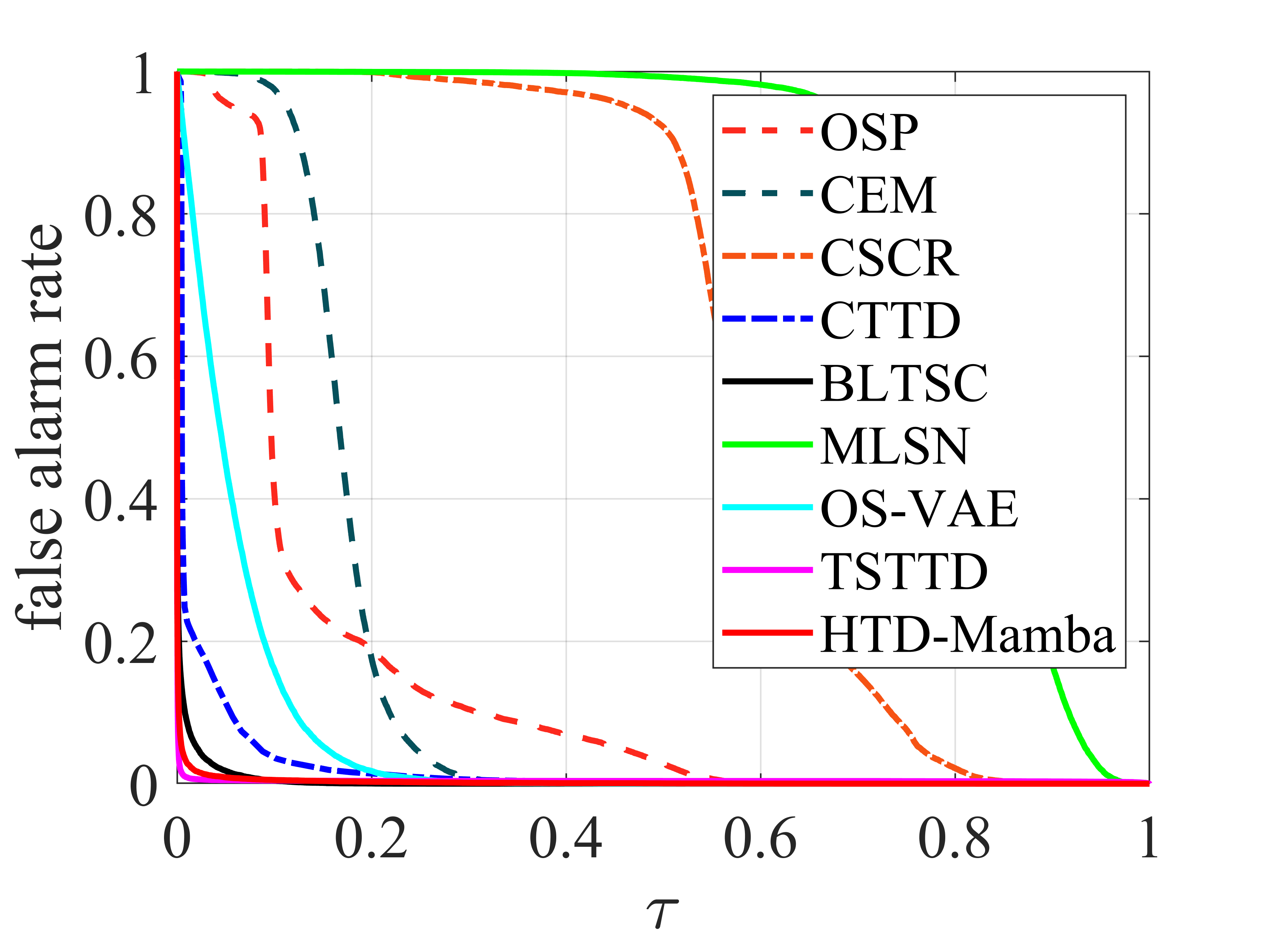}
		}
		\hspace{-18pt}
		\subfigure{
			\includegraphics[width=0.25\textwidth]{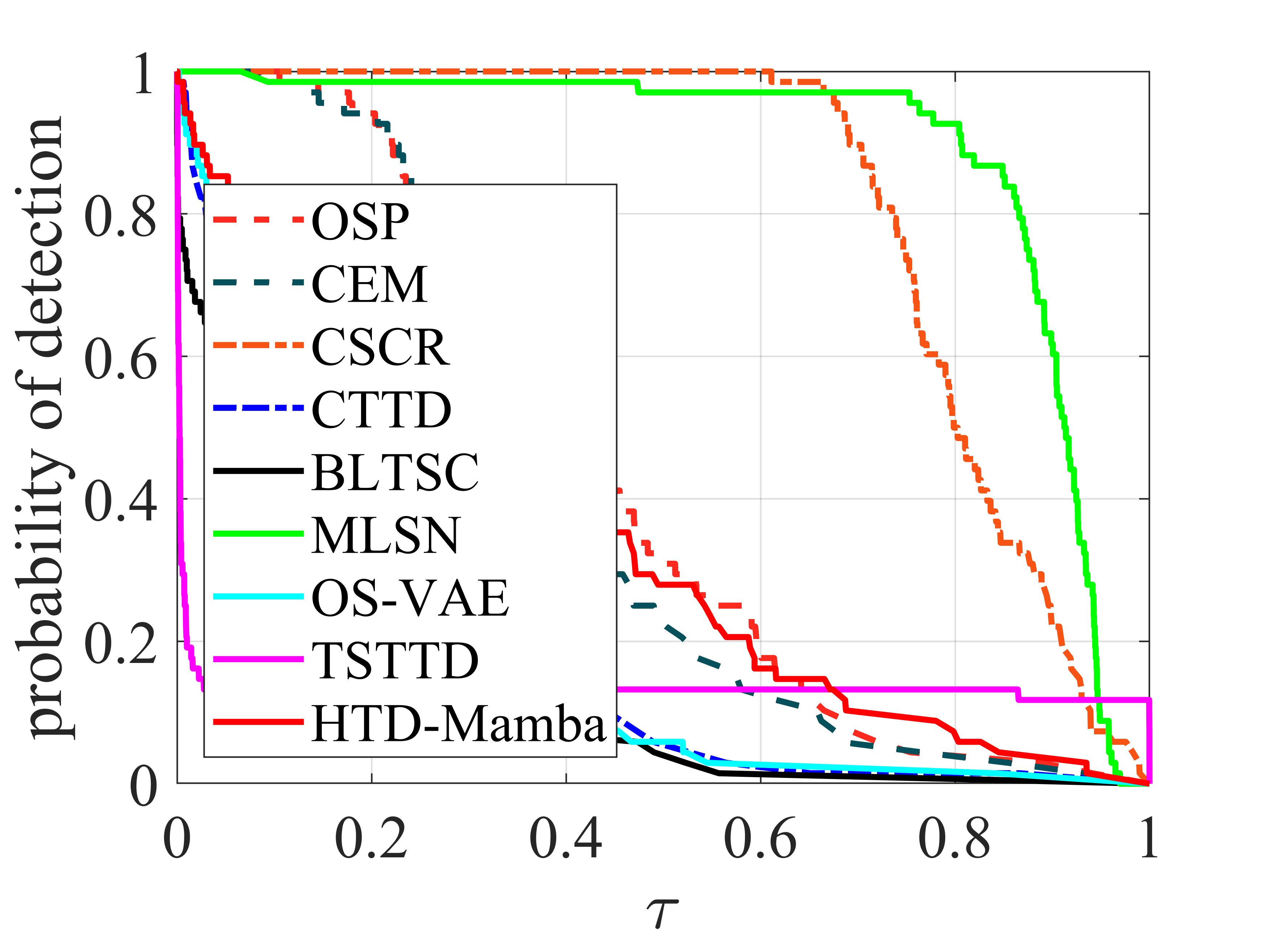}
		}\\
		\parbox{1.0\textwidth}{\centering {\footnotesize (d)}}
	\end{minipage}
	\caption{ROC curves of the competing methods on (a) San Diego I, (b) San Diego II, (c) Los Angeles, and (d) Pavia.  From left to right: 3D ROC curve, 2D ROC curve of $(P_f, P_d)$, 2D ROC curve of $(\tau, P_f)$, and 2D ROC curve of $(\tau, P_d)$.}
	\label{roc_curves_fig}
\end{figure*}
\begin{figure*}[!t]
	\centering
	\subfigure[]{
		\includegraphics[width=0.25\textwidth]{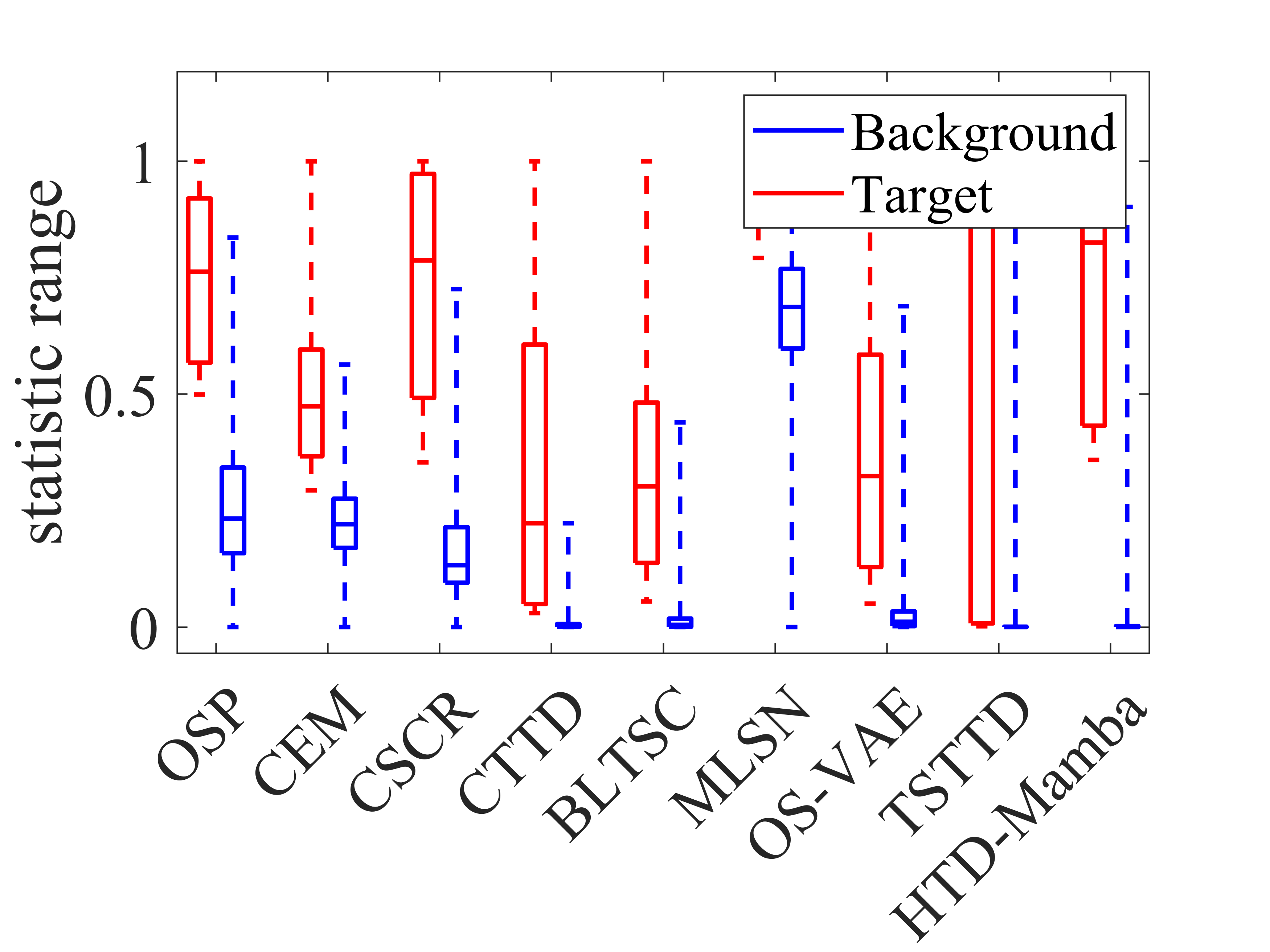}
	}
	\hspace{-18pt}
	\subfigure[]{
		\includegraphics[width=0.25\textwidth]{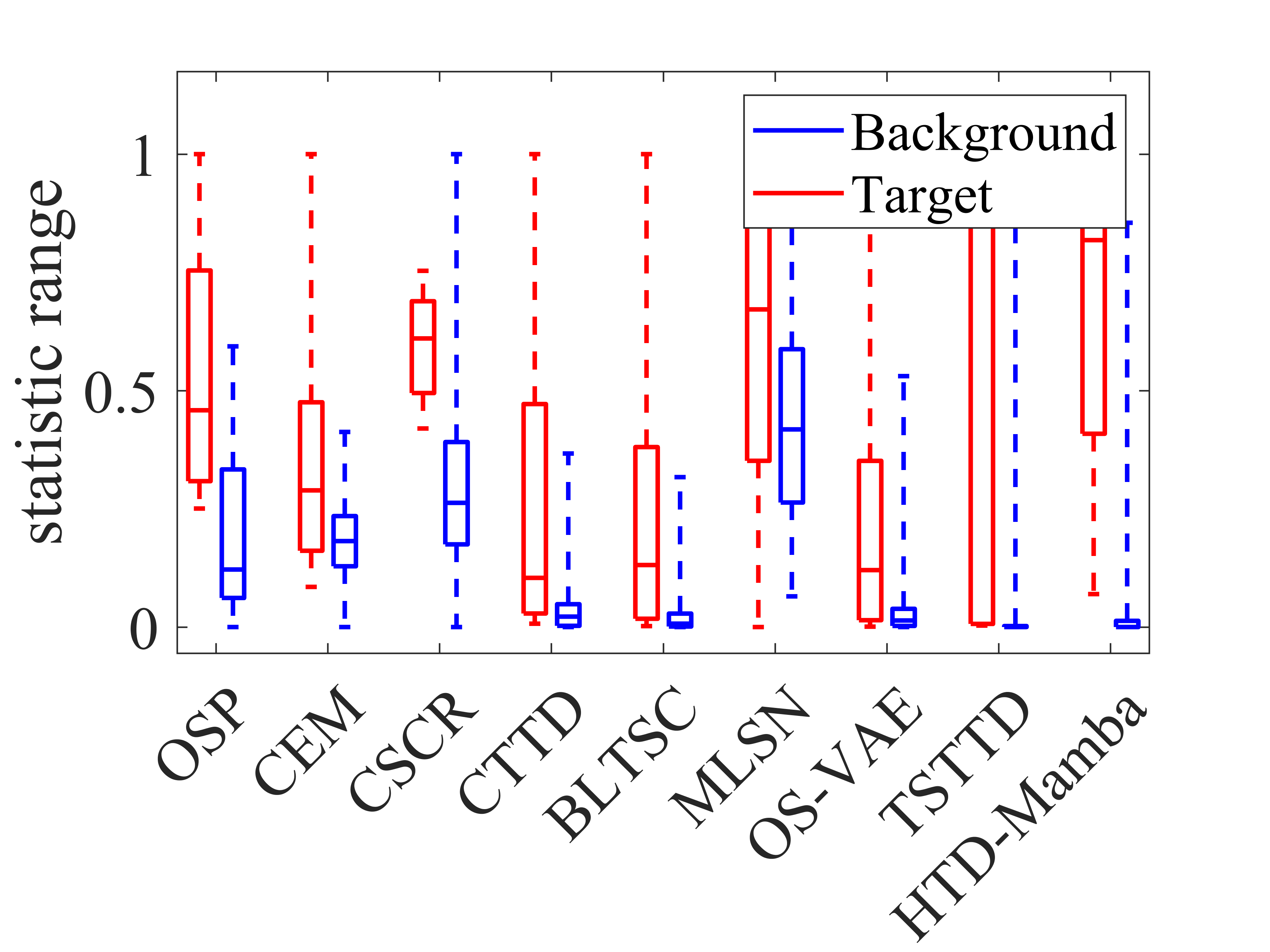}
	}
	\hspace{-18pt}
	\subfigure[]{
		\includegraphics[width=0.25\textwidth]{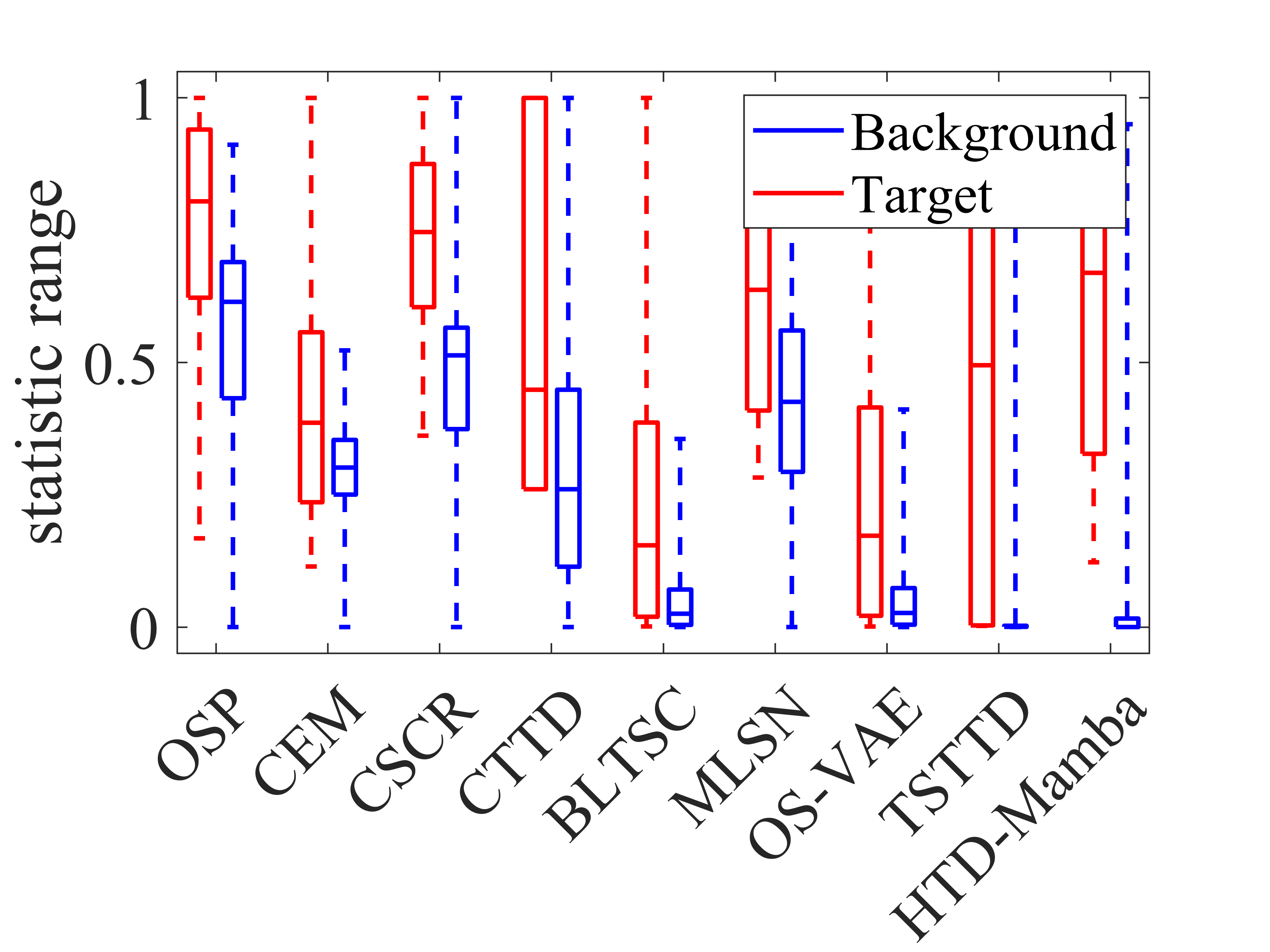}
	}
	\hspace{-18pt}
	\subfigure[]{
		\includegraphics[width=0.25\textwidth]{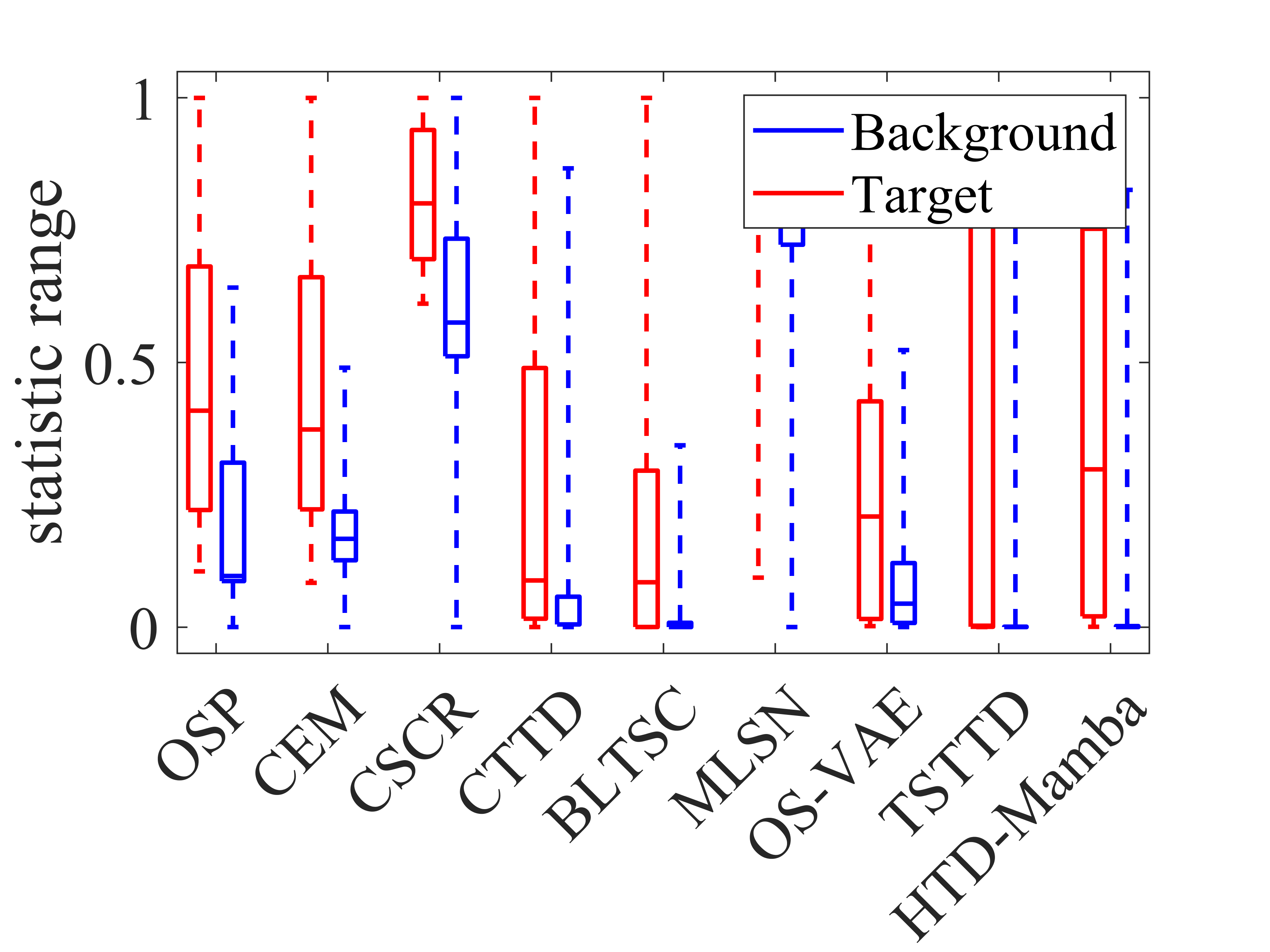}
	}
	\caption{Background-target separation diagrams of competing methods on four datasets. (a) San Diego I. (b)  San Diego II. (c) Los Angeles. (d) Pavia.}
	\label{background_target_separation}
\end{figure*}
\begin{table*}[!t]
	\centering
	\caption{AUC SCORES AND RUNNING TIME (IN SECONDS) OF COMPETING METHODS ON THE FOUR DATASRTS. BOLD REPRESENTS THE BEST RESULT WHILE UNDERLINED THE SECOND.}
	\scriptsize
	\tabcolsep = 9pt
	\renewcommand\arraystretch{1.5}
	\begin{tabular}{clccccccccc}
		\toprule
		Dataset&&OSP&CEM&CSCR&CTTD&BLTSC&MLSN&OS-VAE&TSTTD&HTD-Mamba\\ 
		\bottomrule
		\multirow{5}*{San Diego I}&$\rm{AUC}{(P_f, P_d)}$&0.9972&0.9977&0.9985&0.9982&\underline{0.9987}&0.9940&0.9978&0.9971&\bf{0.9998}\\
		&$\rm{AUC}{(\tau, P_f)}$&0.2461&0.2224&0.1525&0.0055&0.0091&0.6862&0.0163&\bf{0.0007}&\underline{0.0036}\\
		&$\rm{AUC}{(\tau, P_d)}$&0.7616&0.4860&0.7634&0.3461&0.3285&\bf{0.9335}&0.3637&0.7295&\underline{0.7817}\\
		&$\rm{AUC}_{\rm{OA}}$&1.5127&1.2613&1.6094&1.3388&1.3181&1.2413&1.3452&\underline{1.7259}&\bf{1.7779}\\
		&$\rm{AUC}_{\rm{SNPR}}$&3.0947&2.1853&5.0059&62.9273&36.0989&1.3604&22.3129&\bf{1042.1429}&\underline{217.1389}\\
		&Running time&0.4021&\bf{0.3320}&3.6459&\underline{0.3480}&2.4912&12.6212&2.4184&2.7609&1.6895\\
		\bottomrule
		\multirow{5}*{San Diego II}&$\rm{AUC}{(P_f, P_d)}$&0.9599&0.8468&0.9952&0.9220&0.9385&0.7944&0.8875&\underline{0.9969}&\bf{0.9989}\\
		&$\rm{AUC}{(\tau, P_f)}$&0.1672&0.1818&0.2753&0.0286&0.0121&0.4179&0.0183&\bf{0.0018}&\underline{0.0079}\\
		&$\rm{AUC}{(\tau, P_d)}$&0.5156&0.3136&0.6019&0.2263&0.1706&\underline{0.6528}&0.1623&0.6316&\bf{0.7635}\\
		&$\rm{AUC}_{\rm{OA}}$&1.3083&0.9786&1.3218&1.1197&1.0970&1.0293&1.0315&\underline{1.6267}&\bf{1.7545}\\
		&$\rm{AUC}_{\rm{SNPR}}$&3.0837&1.7250&2.1863&7.9126&14.0992&1.5621&8.8689&\bf{350.8889}&\underline{96.6456}\\
		&Running time&\bf{0.0625}&0.4118&10.0797&\underline{0.2980}&2.4624&12.4940&2.4111&2.4944&1.7474\\
		\bottomrule
		\multirow{5}*{Los Angeles}&$\rm{AUC}{(P_f, P_d)}$&0.8937&0.7588&0.9675&0.9180&0.8696&0.8521&0.8773&\underline{0.9838}&\bf{0.9976}\\
		&$\rm{AUC}{(\tau, P_f)}$&0.5893&0.3019&0.4915&0.3192&0.0330&0.4276&0.0349&\bf{0.0043}&\underline{0.0102}\\
		&$\rm{AUC}{(\tau, P_d)}$&\bf{0.7842}&0.4045&\underline{0.7452}&0.6865&0.2016&0.6290&0.2133&0.4948&0.6552\\
		&$\rm{AUC}_{\rm{OA}}$&1.0886&0.8614&1.2212&1.2853&1.0382&1.0535&1.0557&\underline{1.4743}&\bf{1.6426}\\
		&$\rm{AUC}_{\rm{SNPR}}$&1.3307&1.3398&1.5162&2.1507&6.1091&1.4710&6.1117&\bf{115.0698}&\underline{64.2353}\\
		&Running time&\bf{0.0591}&0.3349&4.3352&\underline{0.2420}&2.3861&12.6853&2.4097&2.7999&1.7729\\
		\bottomrule
		\multirow{5}*{Pavia}&$\rm{AUC}{(P_f, P_d)}$&0.9239&0.9419&\underline{0.9489}&0.9049&0.8738&0.7867&0.7823&0.9468&\bf{0.9923}\\
		&$\rm{AUC}{(\tau, P_f)}$&0.1457&0.1704&0.5936&0.0233&\underline{0.0038}&0.8239&0.0567&0.0042&\bf{0.0025}\\
		&$\rm{AUC}{(\tau, P_d)}$&0.4331&0.4089&\underline{0.8165}&0.1655&0.1359&\bf{0.8848}&0.2181&0.1337&0.3481\\
		&$\rm{AUC}_{\rm{OA}}$&\underline{1.2113}&1.1804&1.1718&1.0471&1.0059&0.8476&0.9437&1.0763&\bf{1.3379}\\
		&$\rm{AUC}_{\rm{SNPR}}$&2.9725&2.3996&1.3755&7.1030&\underline{35.7632}&1.0739&3.8466&31.8333&\bf{139.2400}\\
		&Running time&\underline{0.0996}&\bf{0.0883}&6.7446&0.5490&2.6024&26.3778&1.4024&2.8134&2.7239\\
		\bottomrule
	\end{tabular}
	\label{auc_results}
\end{table*}
\subsection{Parameter Analysis}
In this part, we discuss the effects of several key hyperparameters, namely the patch size $p$, the spectral group length $m$, the embedding size $N$, and the network depth $l$.
\subsubsection{Patch Size $p$}
In the SESA process, the transformed view of the center pixel is affected by the patch size. To explore its impact on detection quality, we conducted experiments by varying $p$ within the range of [3, 19], with $p$ being an odd number. The experimental results for the four datasets are shown in Fig. \ref{parameter_analysis}(a). It can be observed that the sensitivity to $p$ varies across different datasets. The performance is relatively stable for San Diego I, shows some fluctuations for San Diego II and Los Angeles, and exhibits the greatest variability for Pavia. This variation is related to the size and distribution of the targets in the different scenes. Considering both detection performance and computational cost, the selected $p$ values for the four datasets are 11, 13, 11, and 5, respectively.
\subsubsection{Spectral Group Length $m$}
The spectral group length $m$ determines the length of the spectral sequence. Smaller $m$ values result in more groups and longer token sequences, while larger $m$ values yield shorter token sequences. The sequence length can significantly impact spectral feature extraction, thereby affecting detection quality. To select an appropriate group length, we varied $m$ within the range of [5, 40], with a step size of 5. The experimental results for the four datasets are shown in Fig. \ref{parameter_analysis}(b). As observed from the figure, the AUC values fluctuate within a relatively small range for the San Diego I, San Diego II, and Los Angeles datasets as $m$ varies. However, the Pavia dataset exhibits significant fluctuations, likely due to large changes in local spectral characteristics. Considering the diversity of the formed spectral sequences and computational cost, the selected $m$ values for the four datasets are 30, 5, 5, and 15, respectively.
\subsubsection{Embedding Size $N$}
The embedding size $N$ represents the dimensionality of the spectral tokens, which also influences detection results. To select an appropriate $N$ value, we varied it within the range of [8, 256] in powers of 2. The experimental results are shown in Fig. \ref{parameter_analysis}(c). As observed, for all four datasets, detection accuracy initially increases with increasing $N$, but then stabilizes or decreases. This is because increasing the feature dimensionality within a certain range enhances the feature representation capability, but excessively large feature dimensions can lead to overfitting. To balance detection accuracy and computational complexity, we set $N$ as 16 for all four datasets.
\subsubsection{Network Depth $l$}
The network depth $l$ represents the number of deep layers in the pyramid SSM. To investigate its impact on detection performance, we compared the AUC scores for network depths ranging from 1 to 5. The experimental results are shown in Fig. \ref{parameter_analysis}(d). As shown in the figure, detection accuracy remains stable or slightly decreases with increasing $l$. This is because the pyramid SSM inherently possesses the ability to extract global spectral features with multiresolution using shallow layers. When the number of layers increases, the model may overfit, leading to decreased accuracy and increased computational complexity. Therefore, we set $l$ to 1 for all datasets to ensure a lightweight model while maintaining detection effectiveness.
\subsection{Performance Comparison}
In this work, we select eight representative advanced methods for comparison: orthogonal subspace projection (OSP) \cite{chang2005orthogonal}, constrained energy minimization (CEM) \cite{du2003comparative}, combined sparse and collaborative representation (CSCR)\footnote[2]{\url{https://fdss.bit.edu.cn/pub/fsyxhyxtktz/yjdw/js/b153191.htm}} \cite{li2015combined}, chessboard-shaped topology for HTD (CTTD)\footnote[3]{\url{https://github.com/sxt1996/CTTD}} \cite{sun2023information}, background learning based on target suppression constraint (BLTSC)\footnote[4]{\url{https://github.com/zhangxin-xd/BLTSC}} \cite{li2015combined}, meta-learning based HTD using a Siamese network (MLSN)\footnote[5]{\url{https://github.com/YuleiWang1/MLSN}} \cite{wang2022meta}, orthogonal subspace-guided variational autoencoder (OS-VAE)\footnote[6]{\url{https://github.com/CX-He/OS-VAE}} \cite{tian2024hyperspectral}, and triplet spectralwise transformer-based target detector (TSTTD)\footnote[7]{\url{https://github.com/shendb2022/TSTTD}} \cite{jiao2023triplet}. Among these, CEM and OSP are statistical-based methods, CSCR is a representation-based method, CTTD is a topological method, and BLTSC, MLSN, OS-VAE, and TSTTD are deep learning-based methods.

For a fair comparison, the key parameters of the competing methods are tuned to optimal levels. Specifically, for the OSP detector, the automatic target generation process (ATGP) \cite{ren2003automatic} is applied to obtain the background subspace, where the number of bases is selected as 22, 2, 1, and 1 for San Diego I, San Diego II, Los Angeles, and Pavia, respectively. In CSCR, the dual window size $(w_{out}, w_{in})$ is set to $(13, 11)$, $(15, 3)$, $(15, 13)$, and $(15, 13)$ for the four datasets, respectively. Additionally, the two regularization parameters $(\lambda, \beta)$ are set to $(0.1, 1)$ for all datasets. For CTTD, the vertical division for spatial disassembly $ver$ and the horizontal division for spectral disassembly $hor$, are set to $(18, 2)$, $(14, 4)$, $(3, 4)$, and $(18, 7)$ for the four datasets, respectively. For BLTSC, the binarization threshold $k$ in coarse detection is fixed at 0.15, the learning rate is $10^{-4}$, the batch size is 64, the number of epochs is 500, and the parameter $\lambda$ in nonlinear background suppression is set to $10^{3}$, $10^{4}$, $10^{4}$, and 10 for the four datasets, respectively. For MLSN, the batch size, learning rate, and epoch are set to 128, $10^{-3}$, and 50, respectively. In the process of guided image filtering, the local window radius is set to 2, and the penalty value is set to 0.04 for all datasets. For OS-VAE, the number of hidden nodes, the network depth, the regularization parameter $\rho$, and the background suppression parameter $\alpha$ are respectively set to 30, 2, $10^{-5}$, and 0.1 for all datasets. For TSTTD, the mixture ratio of background in synthesizing target samples is randomly set between 0 and 0.1. During the training process, the batch size is set to 64, the learning rate is $10^{-4}$, and the total number of epochs is 20.

For an intuitive comparison, Fig. \ref{detection_results} presents the detection results obtained by all competing methods across four benchmark datasets. It illustrates that the proposed HTD-Mamba outperforms other methods by achieving a promising balance between target highlighting and background suppression. The classical detector CEM exhibits weak performance in accurately identifying targets while suppressing the background. Detectors such as OSP, CSCR, and MLSN can detect most target pixels but suffer from poor background suppression. In contrast, BLTSC, OS-VAE, and TSTTD excel at eliminating background interference at the expense of losing significant target information. The CTTD method performs unsteadily, either missing useful information or producing numerous false alarms. Compared to other methods, HTD-Mamba achieves the most satisfactory results in simultaneously highlighting targets and suppressing the background. The detection maps obtained by HTD-Mamba closely align with the ground truths, effectively preserving the shapes and contours of the objects of interest.

Fig. \ref{roc_curves_fig} presents the ROC curves of the competing methods across four datasets, including a 3D ROC curve and three unfolded 2D ROC curves. For the ROC $(P_f, P_d)$, the closer the curve is to the upper left corner, the better the performance. It can be observed that the proposed HTD-Mamba leads in all four datasets, demonstrating superior detection effectiveness. For the ROC $(\tau, P_f)$, which reflects the quality of background suppression, the nearer the curve is to the lower left corner, the better the performance. Among the competing methods, TSTTD and HTD-Mamba achieve the most satisfactory performance in background suppression. Additionally, the closer the curve is to the upper right corner in the ROC $(\tau, P_d)$ space, the better the detection ability of the methods. Although HTD-Mamba does not outperform CSCR and MLSN on the Pavia dataset, its performance remains competitive. Considering the comprehensive capabilities of target prominence and background suppression, our proposed HTD-Mamba achieves the most promising and reliable detection performance across all four datasets.

The ROC analysis alone cannot accurately evaluate detection quality due to overlapping results from multiple methods. Therefore, Table \ref{auc_results} provides a detailed comparison of the AUC scores obtained by competing methods on the four datasets for quantitative evaluation. The best results are marked in bold, while the second-best are underlined. It is evident that the proposed HTD-Mamba achieves the optimal results for the most important $\rm{AUC}{(P_f, P_d)}$ score across the four datasets, which are approximately 0.0011, 0.0020, 0.0138, and 0.0434 higher than those of the second-best, respectively. This indicates its outstanding performance in detection effectiveness. While OSP and MLSN achieve superior $\rm{AUC}{(\tau, P_d)}$ results on the Los Angeles and Pavia datasets, they perform relatively poorly in terms of background suppression due to high $\rm{AUC}{(\tau, P_f)}$ scores. In contrast, BLTSC and TSTTD obtain satisfactory $\rm{AUC}{(\tau, P_f)}$ results across all four datasets but suffer a severe loss of target information. Our proposed HTD-Mamba maintains a promising balance between target detection and background suppression, due to competitive $\rm{AUC}{(\tau, P_d)}$ and $\rm{AUC}{(\tau, P_f)}$ scores. Additionally, HTD-Mamba also achieves superior $\rm{AUC}_{\rm{OA}}$ and $\rm{AUC}_{\rm{SNPR}}$ results across the four datasets, indicating comprehensive detection performance. Table \ref{auc_results} also presents the running time of the competing methods. It is evident that OSP, CEM, and CTTD have lower computational costs due to their simpler algorithms. Our HTD-Mamba achieves competitive computational speed among deep detectors due to the hardware parallel scanning technique, facilitating practical application. In summary, the quantitative assessment provided by AUC scores and running time demonstrates the outstanding performance of the proposed HTD-Mamba.

To further compare the background-target separation performance of the competing methods, Fig. \ref{background_target_separation} shows the separation diagrams for the four datasets. It is evident that our proposed HTD-Mamba achieves the most satisfactory performance, as there is significant distance and minimal overlap between the target and background boxes across all four datasets. Although BLTSC, TSTTD, and HTD-Mamba all achieve superior background suppression performance with small-sized background boxes, our proposed HTD-Mamba demonstrates a more promising target-background separation. Overall, HTD-Mamba consistently produces satisfactory visual detection maps, achieves outstanding quantitative results, and demonstrates promising target-background separation capabilities across all four datasets.
\subsection{Ablation Analysis}
\label{ablation_analysis}
In this part, we verify and analyze the effectiveness of the key contributions of the proposed method. 
\begin{table}[!t]
	\centering
	\caption{Performance comparison of competing spectral augmentation methods on four datasets}
	\scriptsize
	\tabcolsep = 8pt
	\renewcommand\arraystretch{1.5}
	\begin{tabular}{c|cccc}
		\hline
		Method&San Diego I&San Diego II&Los Angeles&Pavia\\ 
		\hline
		Self-Similarity&0.9873&0.9425&0.8544&0.8563\\
		Dropout&0.9917&0.9899&0.9668&0.8868\\
		Gaussian Noise&0.9915&0.9758&0.9549&0.8966\\
		VAE&0.9857&0.9656&0.8886&0.9129\\
		AAE&0.9979&0.9880&0.9317&0.8863\\
		SESA&\bf{0.9998}&\bf{0.9989}&\bf{0.9976}&\bf{0.9923}\\
		\hline
	\end{tabular}
	\label{ablation_SESA}
\end{table}
\begin{table}[!t]
	\centering
	\caption{Effectiveness of pyramid SSM on four datasets}
	\scriptsize
	\tabcolsep = 8pt
	\renewcommand\arraystretch{1.5}
	\begin{tabular}{c|cccc}
		\hline
		Pyramid SSM&San Diego I&San Diego II&Los Angeles&Pavia\\ 
		\hline
		Without&0.9986&0.9978&0.9853&0.9127\\
		With (Level=1)&0.9991&0.9979&0.9921&0.9336\\
		With (Level=2)&0.9991&0.9972&0.9938&0.9824\\
		With (Level=3)&0.9995&0.9985&0.9950&0.9859\\
		With (Level=4)&\bf{0.9998}&\bf{0.9989}&\bf{0.9976}&\bf{0.9923}\\
		\hline
	\end{tabular}
	\label{effect_pyramid}
\end{table}
	\begin{table*}[!t]
		\centering
		\caption{Performance comparison of different backbones on four datasets.}
		\scriptsize
		\tabcolsep = 9pt
		\renewcommand\arraystretch{1.5}
		\begin{tabular}{clccccccccc}
			\toprule
			Dataset&&MLP&LSTM&ResNet&ViT&Mamba&FPN&U-Net&PVT&Pyramid SSM\\ 
			\bottomrule
			\multirow{5}*{San Diego I}
			&Depth&17&2&9&8&11&1&1&1&1\\
			&Params&0.33M&0.32M&0.34M&0.35M&0.34M&0.33M&0.35M&0.34M&0.33M\\
			&FLOPs&0.52G&0.30G&0.45G&0.56G&0.55G&\bf{0.09G}&0.18G&0.11G&0.14G\\
			&$\rm{AUC}{(P_f, P_d)}$&0.9996&0.9993&0.9993&0.9993&0.9996&0.9995&0.9993&0.9990&\bf{0.9998}\\
			&Running time&1.2545&1.1646&1.2236&1.4968&2.6463&\bf{1.1246}&1.3964&1.7053&1.6895\\
			\bottomrule
			\multirow{5}*{San Diego II}
			&Depth&7&1&7&3&4&1&1&1&1\\
			&Params&0.50M&0.49M&0.49M&0.49M&0.49M&0.49M&0.46M&0.51M&0.49M\\
			&FLOPs&1.74G&\bf{0.18G}&0.99G&1.78G&1.62G&0.21G&0.44G&1.03G&1.17G\\
			&$\rm{AUC}{(P_f, P_d)}$&0.9968&0.9983&0.9982&0.9988&0.9978&0.9981&0.9973&0.9977&\bf{0.9989}\\
			&Running time&1.3318&1.1476&1.1841&1.5923&2.3258&\bf{1.1304}&1.3647&1.7536&1.7474\\
			\bottomrule
			\multirow{5}*{Los Angeles}
			&Depth&6&1&5&3&4&1&1&1&1\\
			&Params&0.51M&0.53M&0.51M&0.53M&0.52M&0.53M&0.49M&0.53M&0.51M\\
			&FLOPs&1.63G&\bf{0.20G}&0.82G&1.94G&1.76G&0.23G&0.47G&1.13G&1.27G\\
			&$\rm{AUC}{(P_f, P_d)}$&0.9962&0.9909&0.9963&0.9963&0.9932&0.9961&0.9946&0.9948&\bf{0.9976}\\
			&Running time&1.3216&1.1711&1.1715&1.6874&2.3912&\bf{1.1491}&1.4144&1.7761&1.7729\\
			\bottomrule
			\multirow{5}*{Pavia}
			&Depth&16&2&8&7&10&1&1&1&1\\
			&Params&0.33M&0.37M&0.34M&0.32M&0.33M&0.36M&0.36M&0.35M&0.33M\\
			&FLOPs&0.64G&0.39G&0.53G&0.64G&0.65G&\bf{0.12G}&0.23G&0.15G&0.19G\\
			&$\rm{AUC}{(P_f, P_d)}$&0.9770&0.9267&0.9875&0.9657&0.9506&0.9735&0.9901&0.9818&\bf{0.9923}\\
			&Running time&1.6198&1.4063&1.5302&2.0458&4.1172&\bf{1.3485}&2.0392&2.6473&2.7239\\
			\bottomrule
		\end{tabular}
		\label{priority_pyramid_SSM}
	\end{table*}
	\subsubsection{Effectiveness of SESA}
	The proposed SESA technique encodes surrounding contextual information to create a new spectral view of the center pixel, providing sufficient view pairs for contrastive learning. To verify its effectiveness, we compare SESA with the no-transform strategy (referred to as self-similarity) and several spectral data augmentation methods, including Dropout\cite{srivastava2014dropout}, Gaussian noise, variational autoencoder (VAE)\cite{an2015variational}, and adversarial autoencoder (AAE)\cite{makhzani2015adversarial}. Specifically, self-similarity treats each pixel and its copy as a positive pair, while different pixels are treated as negative pairs. Dropout randomly deactivates a certain percentage (0.5 in this paper) of the bands of each pixel to create a new spectral view, which is then paired with the original pixel to construct the positive sample. The Gaussian noise method adds random noise (20 dB in this paper) to the pixel to generate a degraded view simulating complex scenes. Both VAE and AAE are autoregressive methods that encode the spectral information into a latent space, either by maximizing the variational lower bound or through adversarial training. Pixels can be paired with their regressed variants generated by VAE or AAE to construct view pairs for training. We adopt the same autoencoder architecture used in \cite{wang2023self} for both VAE and AAE, which includes a spectral residual channel attention module and a simple discriminator composed of two FC layers for AAE. For a fair comparison, the detected pixel is directly paired with the target spectrum during inference. 
	
	Table \ref{ablation_SESA} shows the compared ${\rm AUC}(P_f, P_d)$ scores on the four datasets. It can be observed that SESA outperforms other competing data augmentation methods. The self-similarity method cannot effectively recognize similar instances because the exact match criterion is too strict. Both AAE and VAE are parametric and require pretraining. Moreover, these competing methods do not consider spatial contextual information. In contrast, the proposed SESA not only makes full use of spatial information but is also computationally efficient without additional learnable parameters. Therefore, using SESA can efficiently create effective spectral views and significantly improve detection performance.
	\subsubsection{Effectiveness of Pyramid SSM}
	The proposed pyramid SSM can capture multiresolution spectral long-range dependencies to enhance feature representation. To verify its effectiveness, we first compare it with the base model without the pyramid SSM and with models using different pyramid levels. Table \ref{effect_pyramid} shows the comparison results. It can be observed that after removing the pyramid SSM, the detection performance significantly declines, illustrating its effectiveness. Meanwhile, as the level increases from 1 to 4, the ${\rm AUC}(P_f, P_d)$ scores show an overall upward trend. This demonstrates that the multiresolution spectral features can effectively enhance feature representation and improve detection performance. 
	
	To further verify the superiority of the pyramid SSM, we compare it with several mainstream backbones, including MLP \cite{popescu2009multilayer}, long short-term memory (LSTM) \cite{fischer2018deep}, residual network (ResNet) \cite{he2016deep}, vision Transformer (ViT) \cite{dosovitskiy2020image}, Mamba \cite{gu2023mamba}, feature pyramid network (FPN) \cite{lin2017feature}, U-Net \cite{ronneberger2015u}, and pyramid vision Transformer (PVT) \cite{wang2021pyramid}. Among these methods, MLP, LSTM, ResNet, ViT, and Mamba are single-scale based models, while FPN, UNet, and PVT are multi-scale based models considering different resolutions. Since ResNet, ViT, FPN, UNet, and PVT are originally based on RGB images, we modified them into the 1D form for spectral processing. For a fair comparison, the parameter capacity of all the competing models is set to roughly the same by adjusting the network depth. The comparison results are shown in Table \ref{priority_pyramid_SSM}. It can be found that all the competing models achieve promising performance, but our pyramid SSM obtains the best results on all four datasets. Additionally, with similar parameters, pyramid SSM requires relatively low floating point operations (FLOPs), and the running time is also within a feasible range for practical applications. Therefore, the proposed pyramid SSM is effective and efficient for extracting discriminative intrinsic features in HTD due to its consideration of multiresolution global correlations and the hardware parallel scanning mechanism.
	\section{Conclusion}
	This paper introduces a spectrally contrastive learning method based on Mamba for HTD. It addresses the problem of limited training samples by constructing spectral view pairs for contrastive learning and tackles spectral variation by extracting discriminative intrinsic features using a pyramid SSM. Specifically, pixel-level instances are recognized by maximizing the feature similarity of positive view pairs while minimizing that of negative ones. First, we propose an SESA technique to create spectral views while leveraging spatial information, thereby obtaining sufficient view pairs. Then, we transform the view pairs into spectral sequences using group-wise spectral embedding and introduce Mamba to extract global features by capturing long-range dependencies with linear complexity. Furthermore, we develop a pyramid SSM to obtain multiresolution feature representations by extracting the global correlation of spectral sequences with different resolutions and fusing features from different levels. Experiments verify the effectiveness and superiority of the proposed method in terms of quality evaluation and ablation studies.
	\label{conclusion}
	\section*{Acknowledgement}
	We would like to thank the researchers who kindly shared the experimental data and source codes used in this paper. 
\bibliographystyle{IEEEtran}
\bibliography{HTD_Mamba}
\begin{IEEEbiography}[{\includegraphics[width=1in,height=1.25in,clip,keepaspectratio]{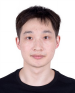}}]{Dunbin Shen} received the B.S. degree in digital media technology from Hubei Minzu University, Enshi, China, in 2016, and the M.S. degree in computer science and technology from Jiangnan University, Wuxi, China, in 2021. He is currently pursuing the Ph.D. degree in signal and information processing with the School of Information and Communication Engineering, Dalian University of Technology, Dalian, China. His research interests include hyperspectral image fusion, deep learning, and hyperspectral target detection.
\end{IEEEbiography}
\begin{IEEEbiography}[{\includegraphics[width=1in,height=1.25in,clip,keepaspectratio]{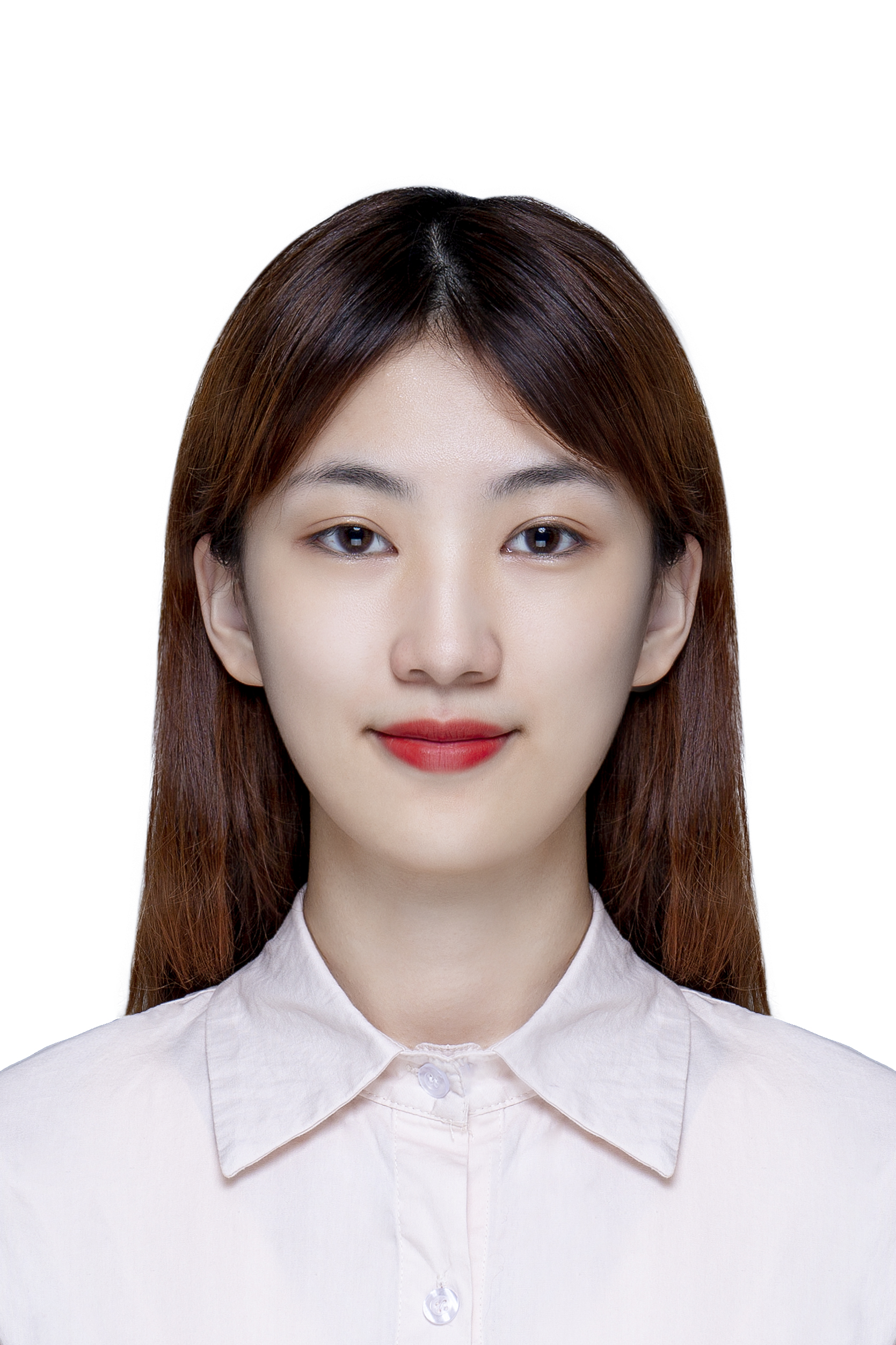}}]{Xuanbing Zhu} received the B.S. degree in internet of things engineering from Dalian Maritime University, Dalian, China, in 2021. She is currently pursuing the Ph.D. degree in signal and information processing with the School of Information and Communication Engineering, Dalian University of Technology, Dalian, China. Her research interests include deep learning, large language model, time series analysis, and data science.
\end{IEEEbiography}
\begin{IEEEbiography}[{\includegraphics[width=1in,height=1.25in,clip,keepaspectratio]{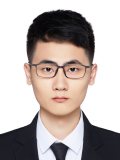}}]{Jiacheng Tian} received the B.S. degree in electronics information science and technology from Dalian Maritime University, Dalian, China, in 2022. He is currently pursuing the M.S. degree in new generation electronic information technology with the School of Information and Communication Engineering, Dalian University of Technology, Dalian, China. His research interests include hyperspectral target detection, and deep learning.
\end{IEEEbiography}
\begin{IEEEbiography}[{\includegraphics[width=1in,height=1.25in,clip,keepaspectratio]{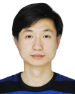}}]{Jianjun Liu}(M'15) received the B.S. degree in applied mathematics and the Ph.D. degree in pattern recognition and intelligence system from Nanjing University of Science and Technology, China, in 2009 and 2014, respectively.\par
	From 2018 to 2020, he was a Postdoctoral Researcher with the Department of Electrical Engineering, City University of Hong Kong, China.
	He is currently an Associate Professor at Jiangnan University.
	His research interests are in the areas of hyperspectral image classification, super-resolution, spectral unmixing, sparse representation, computer vision and pattern recognition.
\end{IEEEbiography}
\begin{IEEEbiography}[{\includegraphics[width=1in,height=1.25in,clip,keepaspectratio]{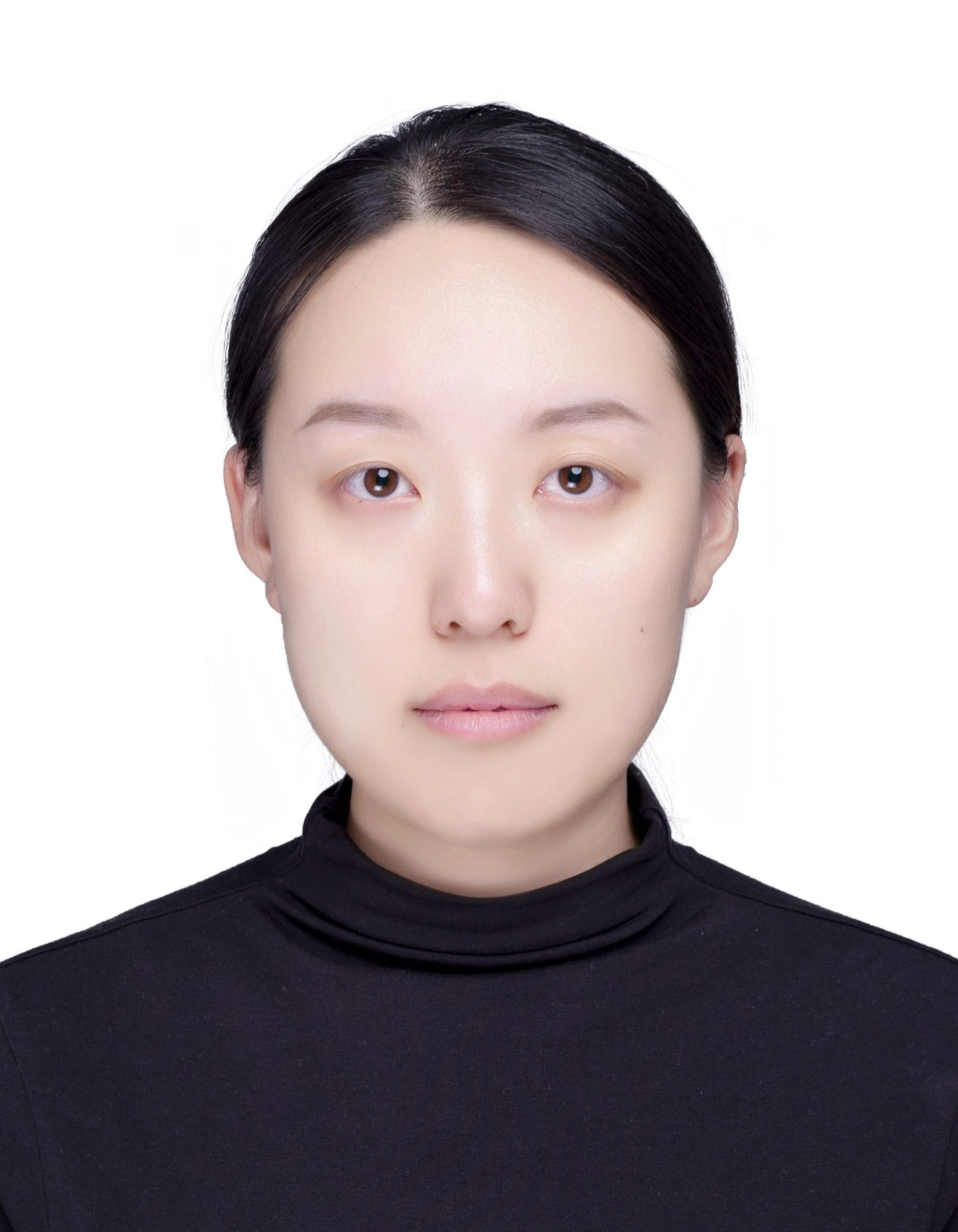}}]{Zhenrong Du}
	received the B.S. degree in geography and Ph.D. degree in agricultural engineering from China Agricultural University, Beijing, China, in 2016 and 2021. She is currently an Associate Professor with the School of Information and Communication Engineering, Dalian University of Technology. Her main research interest is the use of remote sensing techniques to monitor global land use change.
\end{IEEEbiography}
\begin{IEEEbiography}[{\includegraphics[width=1in,height=1.25in,clip,keepaspectratio]{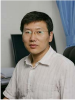}}]{Hongyu Wang}
	received the B.S. and M.S. degrees respectively from Jilin University of Technology and Graduate School of Chinese Academy of Sciences in 1990 and 1993, both in Electronic Engineering. He received the Ph.D. degree in Precision Instrument and Optoelectronics Engineering from Tianjin University, Tianjin, China, in 1997. He is currently a Professor at Dalian University of Technology. His research interests include image processing, image analysis, and remote sensing image classification.
\end{IEEEbiography}
\begin{IEEEbiography}[{\includegraphics[width=1in,height=1.25in,clip,keepaspectratio]{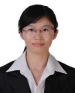}}]{Xiaorui Ma}(M'17)
	received the B.S. degree in applied mathematics from Lanzhou University, Lanzhou, China, in 2008, and Ph.D. degree in communication and information system from Dalian University of Technology, Dalian, China, in 2017. She is currently an Associate Professor at Dalian University of Technology. Her research interests include processing and analysis of remote sensing images, specially hyperspectral image classification and synthetic aperture radar image classification.
\end{IEEEbiography}
\end{document}